\documentclass[pmlr]{jmlr}

\RequirePackage{graphicx}
 \usepackage{booktabs}
 \usepackage{siunitx}

\usepackage{fancyhdr}
\usepackage{hyperref}
\usepackage{bbm}
\usepackage{xcolor}

\usepackage[draft]{fixme}
\fxsetup{inline,nomargin,theme=color}

\usepackage{longtable}
\makeatletter
\def\set@curr@file#1{\def\@curr@file{#1}} 
\makeatother
\usepackage[load-configurations=version-1]{siunitx} 

\newcommand{\Er}{\mathbb{E}}

\newcommand{\Err}[2]{\mathbb{E}_{#2}\left[#1\right]}
\DeclareMathOperator*{\argmax}{arg\,max}
\DeclareMathOperator*{\argmin}{arg\,min}

\theorembodyfont{\upshape}
\theoremheaderfont{\scshape}
\theorempostheader{:}
\theoremsep{\newline}

\jmlrvolume{252}
\jmlryear{2024}
\jmlrworkshop{Machine Learning for Healthcare}

\title[The Data Addition Dilemma]{The Data Addition Dilemma}

\author{\Name{Judy Hanwen Shen}
       \Email{jhshen@stanford.edu}\\ 
       \addr Stanford University
       \AND
       \Name{Inioluwa Deborah Raji}
       \Email{rajiinio@berkeley.edu}\\ 
       \addr UC Berkeley
       \AND
       \Name{Irene Y. Chen}
       \Email{iychen@berkeley.edu}\\ 
       \addr UC Berkeley and UCSF
       }

\begin{document}

\maketitle

\begin{abstract}

In many machine learning for healthcare tasks, standard datasets are constructed by amassing data across many, often fundamentally dissimilar, sources. But when does adding more data help, and when does it hinder progress on desired model outcomes in real-world settings? We identify this situation as
the \textit{Data Addition Dilemma}, demonstrating that adding training data in this multi-source scaling context can at times result in reduced overall accuracy, uncertain fairness outcomes, and reduced worst-subgroup performance. We find that this possibly arises from an empirically observed trade-off between model performance improvements due to data scaling and model deterioration from distribution shift. 
We thus establish baseline strategies for navigating this dilemma, introducing distribution shift heuristics to guide decision-making on which data sources to add in data scaling, in order to yield the expected model performance improvements. 
We conclude with a discussion of the required considerations for data collection and suggestions for studying data composition and scale in the age of increasingly larger models. 
\end{abstract}

\section{Introduction}

Medical institutions will often go to great lengths to increase the size of the available dataset for model development. While some of these data gathering efforts are squarely focused on evaluation ~\citep{dandelion_health}, many are oriented towards compiling larger and larger medical ~\emph{training} sets, composed of data from an increasing number of data sources~\citep{philipsPhilipsIMES, statnewsNeedMuch, ukbiobankAmbitiousProject, ramirez2022all}. Due to the limited data availability in any given data source (e.g., a clinical site), data scaling in such cases involves not just collecting more samples from a single source but also accumulating data across a variety of available sources. This results in data scaling that also incurs changes to training data composition. This leads to models with less predictable outcomes, which can at times perform \emph{worse} than smaller scale in-distribution models.

\begin{figure}
    \centering
    \includegraphics[width=0.9\textwidth]{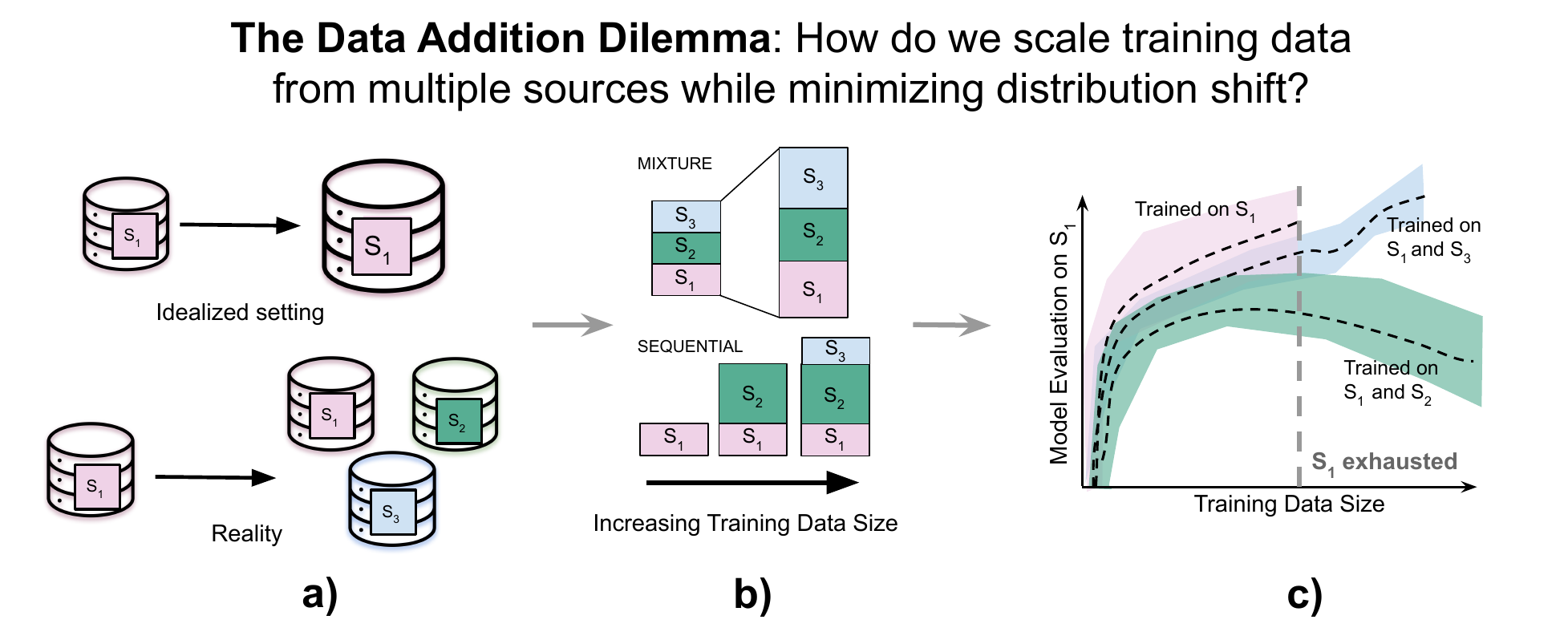}
    \caption{Our work introduces the \textit{Data Addition Dilemma}, where, in a multi-source scaling context, an increase in training dataset size can unexpectedly lead to worse downstream model performance outcomes due to acquired distribution shift.}
    \label{fig:data_dilemma}
\end{figure}

So, when should a practitioner choose to add out-of-distribution data sources to supplement a training set, and which sources should they add such that performance can continue to improve? We identify this data decision-making as part of the \textit{Data Addition Dilemma}, which describes the difficulty to making such choices in the context of multi-source data scaling (Figure~\ref{fig:data_dilemma}).
We take a pragmatic approach to data accumulation and construct scenarios that more explicitly factor in both changes to scale and the corresponding changes to data composition.
In particular, we take a principled approach to formalizing the Data Addition Dilemma, which gives intuition for why increasing the dataset size may not be sufficient to guarantee better performance, especially in the context of dynamic distribution changes. We then test our heuristics for data addition on a real-world dataset that is accumulated from intensive care units across multiple hospitals. 

The focus on increasing the size of available training data reflects a broader perspective on scale common in machine learning settings~\citep{paullada2021data}. The accumulation of data (labeled or unlabeled) in machine learning is largely touted as a reliable solution to many of its modeling problems. The benefits of more data on performance have been observed across many domains including tabular~\citep{viering2022shape}, language~\citep{brown2020language}, vision ~\citep{chen2020simple}, and multi-modal data~\citep{wang2021simvlm}.
Beyond accuracy, increasing dataset size has also been shown to improve adversarial robustness~\citep{carmon2019unlabeled} and robustness against distribution shift~\citep{miller2021accuracy}. Furthermore, when adding more data also improves subgroup representation, group-level disparities in classification can also be reduced~\citep{rolf2021representation,chen2018my}. 

However, in at least the medical context, it is clear that the accumulation of data does not come without notable costs and constraints. Efforts at cross-institutional data collaborations comes only after years of planning, requiring everything from federated data distribution schemes~\citep{rieke2020future, antunes2022federated} and complex standardization initiatives~\citep{hripcsak2015observational,bodenreider2004unified}.
Ultimately, acquiring more data for training involves much more than a naive increase in the number of training samples---in many practical settings, particularly in the medical domain, it involves a variety of possible data accumulation schemes, the most common of which involves the combination of individual data sources (e.g., hospital sites) ~\citep{ukbiobankAmbitiousProject}. While common knowledge in machine learning engineering practice ~\citep{shankar2022operationalizing}, these challenges still remain relatively under-explored theoretically and empirically in machine learning research.

\subsection*{Generalizable Insights about Machine Learning in the Context of Healthcare}

Our work presents the following generalizable insights for machine learning in the context of healthcare:

\begin{enumerate}
    \item \textbf{Problem formulation of the Data Addition Dilemma:} We motivate and formalize cases of data accumulation from single-source and multi-source settings and discuss how the latter is not well-formalized. We then formally present the Data Addition Dilemma, where data composition changes during scaling in the multi-source setting can result in fairly unpredictable downstream model outcomes. 
    \item \textbf{Exploring data composition changes:} We theoretically demonstrate how data composition changes from multi-source scaling can lead to worse model outcomes. We explore multiple distribution shift measures, and discuss how this distribution shift in single-source and multi-source data addition settings can influence performance patterns. 
    \item \textbf{Strategies for data addition:} We present a simple heuristic to determine when to add more data. 
    We discuss the performance impact of the data composition changes to data scaling in the multi-source setting, and illustrate on hospital case studies from a real-world dataset how to best approach the Data Addition Dilemma, and select compatible additional data sources that are most likely to yield performance improvement. 
\end{enumerate}

Most importantly, we hope for this work will be a critical starting point in formalizing the complex dynamics underlying data decision-making as part of the machine learning process. The details of data practices are often overlooked by the machine learning research community altogether---despite its key role in determining the nature of model outcomes~\citep{paullada2021data}. This work represents a strong starting point for a deeper investigation by the machine learning community into more principle-based foundations of meaningful data practices.




\section{Related Work}

\paragraph{Medical distribution shift and data scaling.} The effect of distribution shift has been well-studied in medical settings~\citep{nestor2019feature,wilson2021electronic,wong2021external,koh2021wilds,daneshjou2022disparities,futoma2020myth,guo2022evaluation,chirra2018empirical,yang2023change}. Prior solutions have focused on methods to address distribution shift including stability evaluation~\citep{subbaswamy2021evaluating}, causal methods~\citep{zhang2023did}, and domain adaptation~\citep{subbaswamy2019preventing}; however, the joint problem of distribution shift and data scaling has only recently been proposed. The closest work to ours,~\citet{compton2023more}, reasons about how additional data may not improve performance and studies the problem through the lens of spurious correlations. We focus instead on practical tools for determining when and how to add training data from multiple sources with distribution shift.

\paragraph{Data scaling laws influence model outcomes}
Learning curves (i.e., ``data scaling laws'') have long established the relationship between a machine learning model's generalization performance and training dataset size~\citep{viering2022shape}. This observation that has  persisted, from the dawn of simple models~\citep{cover1967nearest} to our modern neural networks~\citep{bahri2021explaining}. 
Furthermore, adding more data has been suggested as a way to improve not just model accuracy but also model fairness ~\citep{chen2018my,chen2020ethical}, and robustness outcomes~\citep{carmon2019unlabeled, miller2021accuracy}. However, prior considerations are scoped to an in-distribution data setting, where the training set is of a fixed data composition, and sampled in-distribution with respect to the same population as the target test set.

\paragraph{Data composition influences model outcomes}
The composition of training data, at any size, has been shown to influence model outcomes. Data properties such as data diversity, redundancy, and noise can all contribute to model performance, robustness, fairness, and efficiency~\citep {mitchell2022measuring}. These data properties are typically determined by how the data is collected. For example, ~\citet{rolf2021representation} suggests sampling directly from group-specific distributions in order to improve model performance on certain under-represented subgroups. 

Prior work in domain adaptation and distribution shift has also considered data composition independent of size~\citep{quinonero2008dataset, kouw2018introduction, gulrajani2020search, yang2023change}. However, in these works, training data is defined as a pre-existing, static condition, which is not realistic in a multi-source setting. Furthermore, data composition challenges (i.e., distribution shift) are typically addressed with algorithmic interventions rather than data-centric decision-making~\citep{quinonero2008dataset}. 

\paragraph{Realistic data scaling impacts data composition}

When realistically increasing the size of a training dataset, compositional changes in the data can be introduced. Thus, a more pragmatic perspective to data scaling that factors in changes to overall data composition is required---we call this process \emph{data accumulation} (Figure \ref{fig:data_accumulation}). In surveys, sampling bias exacerbates mis-estimation error as the sample size increases, as observed in settings estimating vaccine uptake~\citep{bradley2021unrepresentative} and election polling~\citep{meng2018statistical}. 

Thus far, relatively few works in machine learning have critically examined the effect of increasing training dataset size while factoring in the potential changes to data induced by scaling. In the image classification setting, recent work found that performance heavily depends on the pre-training source data ~\citep{nguyen2022quality} and spurious correlations may be introduced when combining data sources~\citep{compton2023more}. Using a theoretical model, \citet{hashimoto2021model} looks at data as a fixed mixture of different sources (e.g., different categories of Amazon reviews) to characterize excess loss as dataset size increases. ~\citet{tae2021slice} focus on selectively acquiring data from a fixed set of sources, while maintaining model performance and fairness.

\section{The Data Addition Dilemma}
\label{sec:model}

\begin{figure}[ht]
    \centering  
    \includegraphics[width=0.6\textwidth]{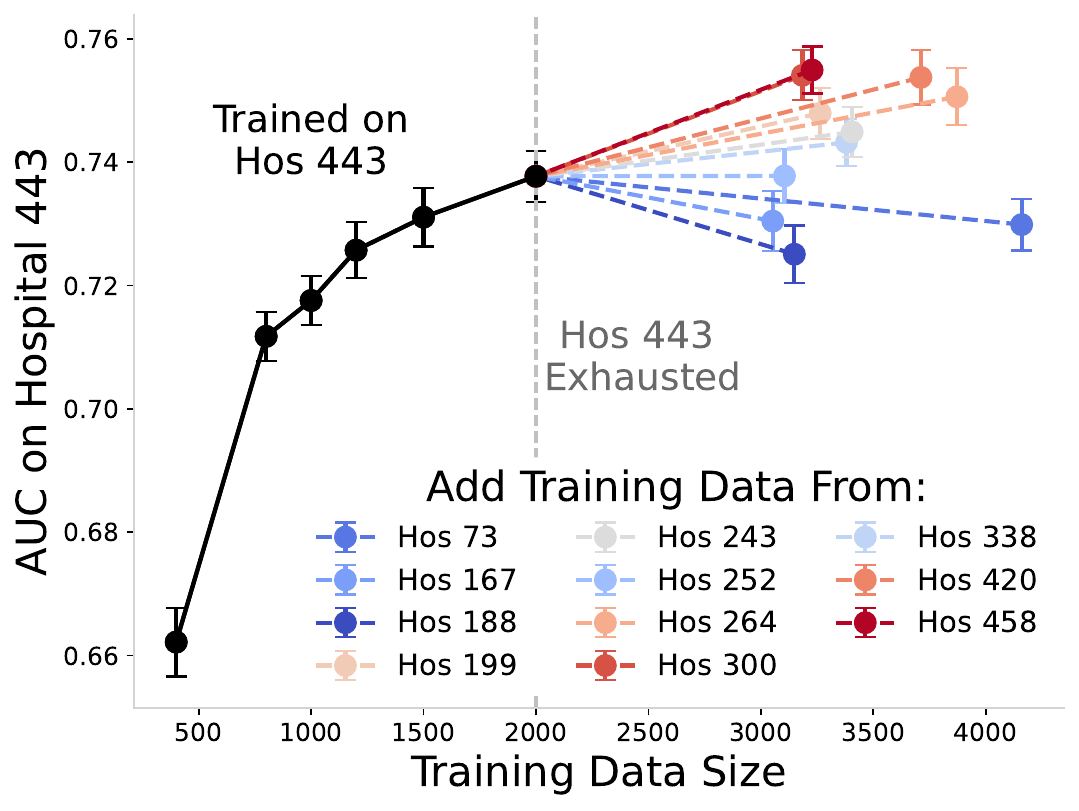}
    \caption{We illustrate the \textit{Data Addition Dilemma}. After hospital 443 has exhausted its in-distribution training data, we can consider adding training data from any of the 11 available hospitals in the eICU dataset. The addition of any available hospital data source can lead to uncertain outcomes and does not guarantee improved test performance for the original setting of hospital 443. 
    }
    \label{fig:illustrative}
\end{figure}


Much of the past work on the impact of data scaling on model performance assumes a \emph{single-source setting}, where data from a single data sampling process is scaled up, and the distribution of the dataset remains fixed. That is, as the number of training samples $n$ increases within a fixed distribution $D_{i}$, then performance on a test set approximately also within that fixed distribution $D_{i}$ generally improves. It is under this setting that many claims on data scaling are typically considered~\citep{viering2022shape}.

It is equally well-studied at this point that distribution shift changes between the training and test set leads to performance drops~\citep{quinonero2008dataset}. 
Increasingly, a set of recent studies ~\citep{meng2018statistical, bradley2021unrepresentative} have extrapolated this result to the setting of out-of-distribution single source scaling; these studies concluded that an increase in the number of training samples $n$ with a distribution $D_{j}$ that is unrepresentative of a test distribution $D_{i}$ may lead to worse model outcomes as $n$ increases and the distribution shift worsens. 

However, in most practical settings, due to the limited availability of data samples in a given source, data accumulation rarely occurs in a single-source setting but rather in a \emph{multi-source setting}, where the final training dataset is pieced together from multiple distinct data sampling processes~\citep{hashimoto2021model,tae2021slice}. We extend this to the consideration of ~\emph{multi-source scaling}, where, in order to practically increase the size of the training dataset, data from multiple sources are collected and combined. Unlike the single source setting, there is not one static scaled-up data collection process. Instead, the resultant larger dataset is a mixture of multiple data sources. This presents more complex data composition changes as the dataset size increases. In other words, $D_{train}$ and $n_{train}$ are dynamically defined by the collective size and composition of the individual sources that compose the overall training set. 

In such a context, it is unclear how model performance is impacted by the addition of sources introduced to increase training dataset size (e.g., Figure~\ref{fig:illustrative}). Practitioners are thus faced with what we define as the ~\emph{Data Addition Dilemma} (Figure ~\ref{fig:data_dilemma})---i.e., will adding data from an additional training data source yield performance improvement or decay?

This \textit{Data Addition Dilemma} is most readily observable in machine learning for healthcare. In healthcare settings, there are many practical reasons to seek additional sources to increase training dataset size: positive label values like disease diagnosis may be underrepresented in the available hospital site, or certain demographic subgroups may also be underrepresented. In many cases, the easiest approach to increasing the available training data is to combine the data from the considered setting with additional samples from other hospitals---a practice that has led to several multi-site data sharing initiatives for not just model evaluation, but pooling data across multiple sites for model ~\emph{training}~\citep{pollard2018eicu}. This common scenario is thus a multi-source training data scaling problem, where the distribution shift incurred by the addition of out-of-distribution data sources has some trade-off with the expected performance improvement of increasing training dataset size, yielding a fairly variable range of possible overall effects.
In this section, we attempt to formalize the dilemma itself and any intuition we can identify in order to inform actionable insights for data addition.

\paragraph{Preliminaries}
Let $\{x, y\}^n \sim \mathcal{D}$ be data generated from some underlying distribution. Let $D_{S_1}, ..., D_{S_m}$ be a series of empirical distributions sampled with varying types of sampling bias from the underlying distribution $\mathcal{D}$; these empirical source distributions are finite and are different fixed sizes $n_{s_1}, ..., n_{s_m}$. The training distribution at a particular training dataset size $n$, $D_{train,n}$, is composed of these $m$ sources. The reference distribution $D_{test}$, where we hope to achieve good performance, is sampled directly from $\mathcal{D}$ without any sampling bias. We define $\delta: \mathcal{P}(\mathcal{X}, \mathcal{Y}) \times \mathcal{P}(\mathcal{X}, \mathcal{Y}) \rightarrow \mathbb{R}_{\ge 0}$ 
as a divergence between distributions. 


To understand how the divergence between train and test data can increase as we enlarge the dataset, we consider the following example.

\begin{example}
Consider a training set of two sources: $D_{S_1}$ a small high-quality dataset relative to $D_{test}$, and $D_{S_2}$ a large lower-quality dataset relative to $D_{test}$. We can model the divergence between train and test distributions as follows \textbf{if} $\delta$ composed linearly:
\begin{align*}
  \delta(D_{train, n}, D_{test})  = \begin{cases}
      \delta(D_{S_1}, D_{test}) \; \text{if $n \le |D_{S_1}|$}\\
      \frac{|D_{S_1}|}{n}\delta(D_{S_1},D_{test}) + 
 (1 - \frac{|D_{S_1}|}{n}) \delta(D_{S_2}, D_{test}) \;  \text{otherwise}\\
    \end{cases}  
\end{align*}
\end{example}
While we cannot assume that divergences compose linearly, we can limit our scope to $f$-divergences and use the convexity of this class of divergences to show that in sequential data addition, $\delta(D_{train, n}, D_{test})$ might increase with $n$ in the general sequential data accumulation setting. 

\paragraph{Sequential Data Addition} 
When adding data sequentially, the training data is a strictly increasing collection of the underlying sources: $\hat{D}_{train, n}= (\bigcup_{i=1}^{k-1} \hat{D}_{S_i}) \cup \frac{n-\sum_{i=1}^{k-1} n_{s_i}}{n_{S_k}}\hat{D}_{S_{k}}$\footnote{$\hat{D}$ denotes the \textit{set} of examples or data points in the distribution D}. Here, $k$ is set to the source index such that $\sum_{i=1}^{k-1} n_{s_i} < n \le \sum_{i=1}^{k} n_{s_i}$. In other words, for a desired dataset size $n$, we start by adding data from the first source and continue to add data from sources sequentially until we reach $n$. This addresses the common scenario where acquiring more data incurs additional cost; all data from existing sources are used before a new source is introduced. The resulting distribution can also be viewed as a mixture of source distributions where $\alpha$ depends on $n$: ${D_{train, n} = \sum_{i=1}^{k} \alpha_i D_{S_i}}$ where $\alpha_i=\frac{n_{s_i}}{n}$ for $i<k$ and $\alpha_m=\frac{n-\sum_{i=1}^{k-1}n_{s_i}}{n_{s_m}}$. 

A key observation we make is that $\delta(D_{train, n}, D_{test})$ in the sequential case will actually depend on the number of samples in the multi-source scaling setting. In the following Lemma, we will show sufficient conditions for when adding an additional data source will increase the resulting divergence between the training and test distributions. 





\begin{lemma}
Let $D_{train, n}$ be constructed by adding data sequentially from $k$ sources: $D_{S_1}, ..., D_{S_k}$, then if $\delta(D_{S_k}, D_{test}) - \frac{cn}{n_{s_k}} \ge \delta(D_{train, n}, D_{test}): $
\[
\delta(D_{train, n}, D_{test}) \ge \delta(D_{train, n-n_{s_k}}, D_{test})
\]
where $\delta$ belongs to the family of $f$-divergences and $c$ is a divergence-dependent constant where $ 
{\delta(D_{train,n}, D_{test}) + c = \sum_{i=1}^{m} \frac{n_{s_i}}{n} \delta(D_{S_i}, D_{test})}$.
\label{lem:div}
\end{lemma}

Lemma \ref{lem:div} gives a relationship between the new data source and the test set that would cause the divergence between train and test distributions to increase with $n$ in the sequential data addition case\footnote{See proof in the appendix}. In a fixed dataset size setting, \citet{acuna2021f} relates increased divergence to empirical risk for $f-$divergences in particular by giving a generalization bound. Prior works have also considered different discrepancy measures including $L_1$ distance~\citep{ben2006analysis}, $\mathcal{H}\Delta\mathcal{H}$ divergence~\citep{ben2010theory}, and margin disparity discrepancy~\citep{zhang2019bridging}. In conjunction with Lemma \ref{lem:div}, these results from prior works give an intuition for a larger empirical risk upper bound when the divergence between train and test distributions increases. 

\paragraph{Practical Strategies For Data Addition} 
In Figure \ref{fig:illustrative}, we illustrate the setting when there are multiple additional data sources available and we have the choice to add one additional hospital to the training data in order to perform better on a test set from the reference hospital. We can denote the performance metric that we care about, e.g., area under the receiver-operator curve (AUC), in terms of train and test datasets as: $AUC(D_{train,n}, D_{test})$. Given a set of available source hospitals ${H_i}$, and a reference hospital of $H_j$, in order to decide which additional hospital data to add, we want to find a hospital $H^*$ to add such that $H^* = \argmax_{H_i}(AUC(D_{H_i} \cup D_{H_j}, D_{H_j}) - AUC(D_{H_j}, D_{H_j}))$ (Section \ref{sec:addition}, Figure~\ref{fig:addition_lr_auc_change}). This represents the performance difference from adding an additional source to the training set, compared to the in-distribution performance under sequential data accumulation.

However, as we observe in Figure \ref{fig:illustrative}, the choice for $H_i$ is not immediately obvious because additional data might in fact lead to performance degradation. In this sequential setting, the increase in overall training data size might also induce a larger distributional difference between the new training set and the reference data (See Theorem \ref{thm:gen}). Furthermore, if we want to add more than 1 additional hospital, the number of combinations of additional sources is exponential, making brute-force computation difficult in practice.

In this paper, we introduce a model of data addition that explains why adding data can increase the distributional distance between train and test data thus increasing the upper bound in expected error. For our problem, we care more about comparing which sources to add rather than estimating the exact divergence or performance. 
We can thus simplify this problem into finding the $\tilde{H}^*$ for the best out-of-distribution (OOD) performance instead, i.e., {$\tilde{H}^* = \argmax_{H_i}(AUC(D_{H_i}, D_{H_j}) - AUC(D_{H_j}, D_{H_j}))$} (Section \ref{sec:shift}, Figure ~\ref{fig:auc_diff}a). However, in reality, we often do not even wish to actually calculate OOD performance differences -- we instead want to decide whether to use the additional data from hospital $i$ before training a full model on this data from hospital $i$. If we assume that all available additional hospitals have a constant size $n$, we can further simplify the problem by searching for an empirical divergence metric that is correlated with OOD performance (i.e., significant $\rho((AUC(D_{H_i},D_{H_j}) - AUC(D_{H_j}, D_{H_j})), \delta(D_{H_i}, D_{H_j}))$) that can be computed with fewer samples. We approximate $\delta(D_{H_i}, D_{H_j})$ using several different heuristics (see Section \ref{sec:heuristic}, Figure~\ref{fig:dist_metrics}).

Using our most correlated heuristic, we can then examine what happens when we add the three best and worst hospitals in terms of divergence from the test hospital. Following the logic from above, we can aim to minimize the total heuristic score over a combination of sequential selections -- for instance, for best three, we can sequentially choose $\tilde{H}^* \approx \argmin_{H_i}(\delta(D_{H_i}, D_{H_j}))$, where $D_{H_j}$ is the reference hospital dataset, and $D_{H_i}$ is any hospital data available to be added to the full training set (see Figure \ref{fig:seq_hos443_lr}). With this case study, we offer one concrete strategy to pragmatically approach the data addition dilemma.

\section{Experiment Setup}
\label{sec:exp}

\subsection{Datasets}
\label{sec:exp-data}

For our investigation, we focused primarily on the eICU Collaborative Research Database~\citep{pollard2018eicu} due to the rich feature sets and open-access availability. The eICU dataset contains 200,859 patient unit encounters in 208 hospitals across the US. We followed the same data cleaning and exclusion criteria for the 24-hour mortality prediction task as \citet{van2023yet} without additional feature generation. We selected the top 12 hospitals, which are all above 2000 patient unit encounters for the task, and set a fixed training set size of 1500 and a test set size of 400 patient encounters for each hospital. We did not perform data imputation in order to best capture the differences in observed data distributions between hospitals. 
In the appendix, we include additional experiments using the MIMIC-IV dataset~\citep{johnson2020mimic},
additional details, and synthetic data experiments\footnote{Code for all experiments can be found at \url{https://github.com/the-chen-lab/data-addition-dilemma}}.

\subsection{Models and Evaluation}
\label{sec:exp-models}
We extended the Yet Another ICU Benchmark package, version 2.0~\citep{van2023yet} for cohort development, training pipeline, and evaluation protocols. We developed custom hospital-based cohort selection and more comprehensive evaluation metrics. Each experiment for each model was performed across 5 validation folds and repeated 5 times (25 models and data splits per experiment). The results we report are averaged first across 5 validation folds and then across the 5 repetition runs; we considered $N=5$ as the 5 repetitions of the entire cross-validation procedure when calculating standard error. Together, we trained over 28,000 models for the main results of our paper.    


We investigated three models with the largest contrast in learning dynamics implemented in YAIB~\citep{van2023yet}: a Logistic Regression (LR)~\citep{cox1958regression} model, a tree-based Light Gradient Boosting Machine (LGBM)~\citep{ke2017lightgbm} model, and a Long Short-Term Memory (LSTM)~\citep{hochreiter1997long} model. We tuned hyperparameters for the LSTM model and use a smaller hidden dimension than used for the full training set. We used the same set of features for all models and do not include additional features for non-deep learning models. Unless otherwise noted (i.e., Figure~\ref{fig:illustrative}), we trained our model on 1500 samples from each available training hospital and compute the metrics on 400 samples from the test hospital. 

Let $f$ denote the model we are evaluating and let $g$ be a group function that maps each data point to a subgroup, we evaluated the following metrics over $D_{test}$: 
\begin{itemize}
    \item\textbf{Area under the Receiver-Operator Curve (AUC)}: 
   $\frac{\sum_{x_0 \in D^0} \sum_{x_1 \in D^1} \mathbbm{1}[f(x_0) < f(x_1)] }{|D^1| |D^0|}$ where  $\mathbbm{1}[f(x_0) < f(x_1)]$ denotes the indicator function which returns 1 if $f(x_0) < f(x_1)$ otherwise return 0. $D^0, D^1$ refer to the set of negative and positive examples, respectively.
\item \textbf{Worst group accuracy}: The accuracy on the worst-performing subgroup and the metric of interest in studying subpopulation shifts in the distribution shift literature~\citep{koh2021wilds} ($\min_{g'}\Er_{(x, y) \sim D_{test}}{f(x) = y |g(x) = g'}$). Due to the instability of comparing AUC across subgroups for varying thresholds~\citep{kallus2019fairness}, we analyzed accuracy with a single fixed threshold chosen across validation datasets for worst group accuracy and disparity.
\item \textbf{Disparity}: The difference between the best and worst-performing subgroups:
$$\max_{g'} \Er_{(x, y) \sim D_{test}}{[f(x) = y |g(x) = g']} - \min_{g'}\Er_{(x, y) \sim D_{test}}{[f(x) = y |g(x) = g']}$$
\end{itemize}

\subsection{Characterizing Distribution Shift}
\label{sec:exp-dist}
The dominant measurement of distribution shift in prior work is the test-time drop in overall and subgroup model performance---both in the clinical setting~\citep{spathis2022looking, schrouff2022diagnosing, guo2023ehr} and the more general machine learning literature~\citep{koh2021wilds, wiles2021fine, yang2023change}. In the electronic health records domain, normative changes to data composition (e.g., subgroup size, outcome distribution) have been characterized \citep{spathis2022looking, schrouff2022diagnosing}. However, it remains unclear how these changes in composition contribute to predictor outcomes when data of differing composition is used. As a baseline, we examined normative characterizations of dataset composition. These normative characteristics include demographic distributions of age, sex, and race and mortality rates of demographic distributions of age, sex, and race. We used the pairwise Euclidean distance between hospitals to compute our normative metrics. 

Another approach to estimating distance is to compute distributional distances directly from features and labels. Recent works in the applied statistics literature have aimed to decompose performance drops due to distribution shifts into different components~\citep{cai2023diagnosing}. The key to such estimations is estimating the densities of both the source and target distributions. However, since clinical records are a complex\footnote{We find that principal component analysis requires $~40$ components to explain $95\%$ of the variance in the eICU Mortality Task}, density estimation is difficult due to the high dimensionality of the data.  For high dimensional data, prior score-based methods estimate the density ratio so that $f$-divergences including KL divergence can be directly computed~\citep{nguyen2007estimating, rhodes2020telescoping, choi2022density}.

Let $P= D_{train, n}$ represent the training distribution and $Q=D_{test}$ represent our reference distribution. Given the standard definition of Kullback-Leibler (KL) divergence $KL(P || Q) = \Er_{x \in P} \left [\log \frac{P(x)}{Q(x)} \right ]$, we wish to estimate the density ratio $r(x) = \frac{P(x)}{Q(x)}$. To estimate $r(x)$, we can derive a predictor $s:\mathcal{X} \rightarrow [0, 1]$ where we give labels 1 to datapoints in $P$ and 0 to datapoints in $Q$. Thus $s_{PQ}(x)$ is fit using the points from both $P$ and $Q$. The resulting ratio is:
\[r(x) = \frac{P(x)}{Q(x)} = \frac{\Pr[x \in P]}{\Pr[x \in Q]} = \frac{s_{PQ}(x)}{1-s_{PQ}(x)}\]
We can then compute the KL divergence between $P$ and $Q$ as: $$KL(P || Q) = \Er_{x \in P} \left [\log r(x) \right ]$$  

Another heuristic we considered is the expected predictor score itself: $\textsc{Score}(s, P, Q) = \Er_{x \in P}[s_{PQ}(x)]$. Intuitively, if $P$ and $Q$ and very similar, the probabilities output by $s$ on the empirical distribution on $P$ should be around $0.5$ or less. If $P$ and $Q$ are very different, $s$ should output values close to $1$ on the empirical distribution of $P$. As an additional heuristic, we can compute the expected predictor probability on $P$ while assuming a uniform probability for the empirical samples of $P$; the resulting predictor score $s_{PQ}$ gives us a score for how likely a sample is to originate from $P$.

There may be changes in the distribution beyond just the covariates shift or different features. For example, hospital 73 has only $1.9\%$ percent mortality compared to 5-10\% in other hospitals. Therefore, we considered the KL divergence between the joint distribution of features and labels between the training and test distributions. We can compute the same set of distances (ratio-based KL and Score) with both the features and the labels. Empirically, we used a logistic classifier~\citep{cai2023diagnosing} to estimate $s$ and clip values of $s$ between $[0.01, 0.99]$ for numerical stability.

In addition to normative distances, we considered the following 4 distance metrics: \begin{itemize}
\itemsep0em
    \item \textsc{KL Ratio X} =  $\Er_{x \in P}[\log \frac{P(x)}{Q(x)}]$ denotes the KL divergence using only the covariates/features.
    \item \textsc{KL Ratio XY} = $\Er_{(x, y) \in P}[\log \frac{P(x,y)}{Q(x,y)}]$ denotes the KL divergence using the covariates/features and labels.
    \item \textsc{Score X} =  $\Er_{x \in P}[s_{PQ}(x)]$ denotes the expectation of the score function $s$ using the covariates/features.
    \item \textsc{Score XY} =  $\Er_{(x, y) \in P}[s_{PQ}(x,y)]$ denotes the expectation of the score function $s$ using the covariates/features and the labels.
\end{itemize}

\section{Results and Analysis}

We present our analysis of case studies of the Data Addition Dilemma, and characterize a possible heuristic to guide data decision-making for training data composition while scaling.

\subsection{Performance Patterns Under Single-Source Distribution Shift}
\label{sec:shift}
\begin{figure}[ht]
    \centering
\subfigure[]{\label{fig:auc_diff1}\includegraphics[width=65mm]{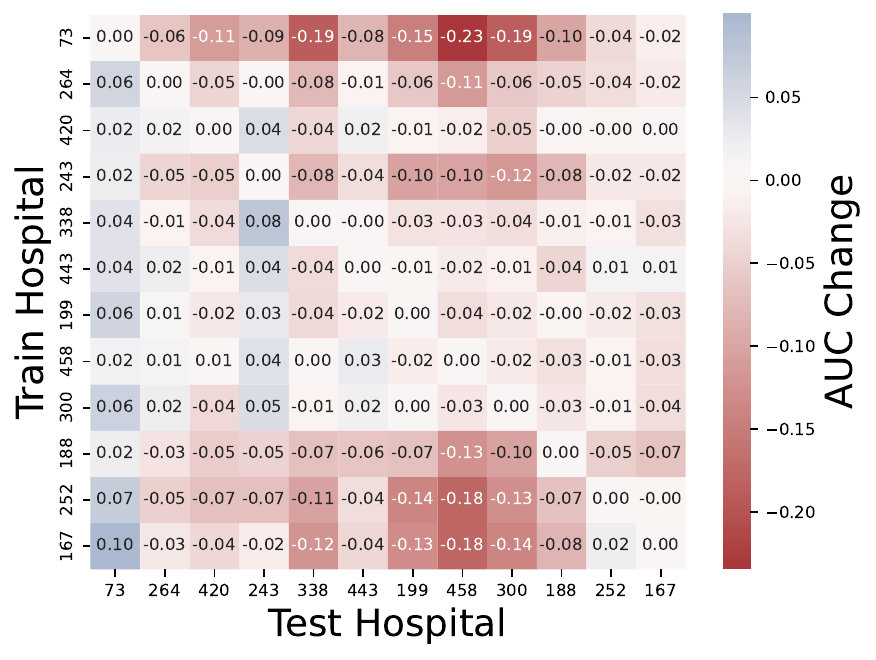}}
\subfigure[]{\label{fig:lr_lstm1}\includegraphics[width=65mm]{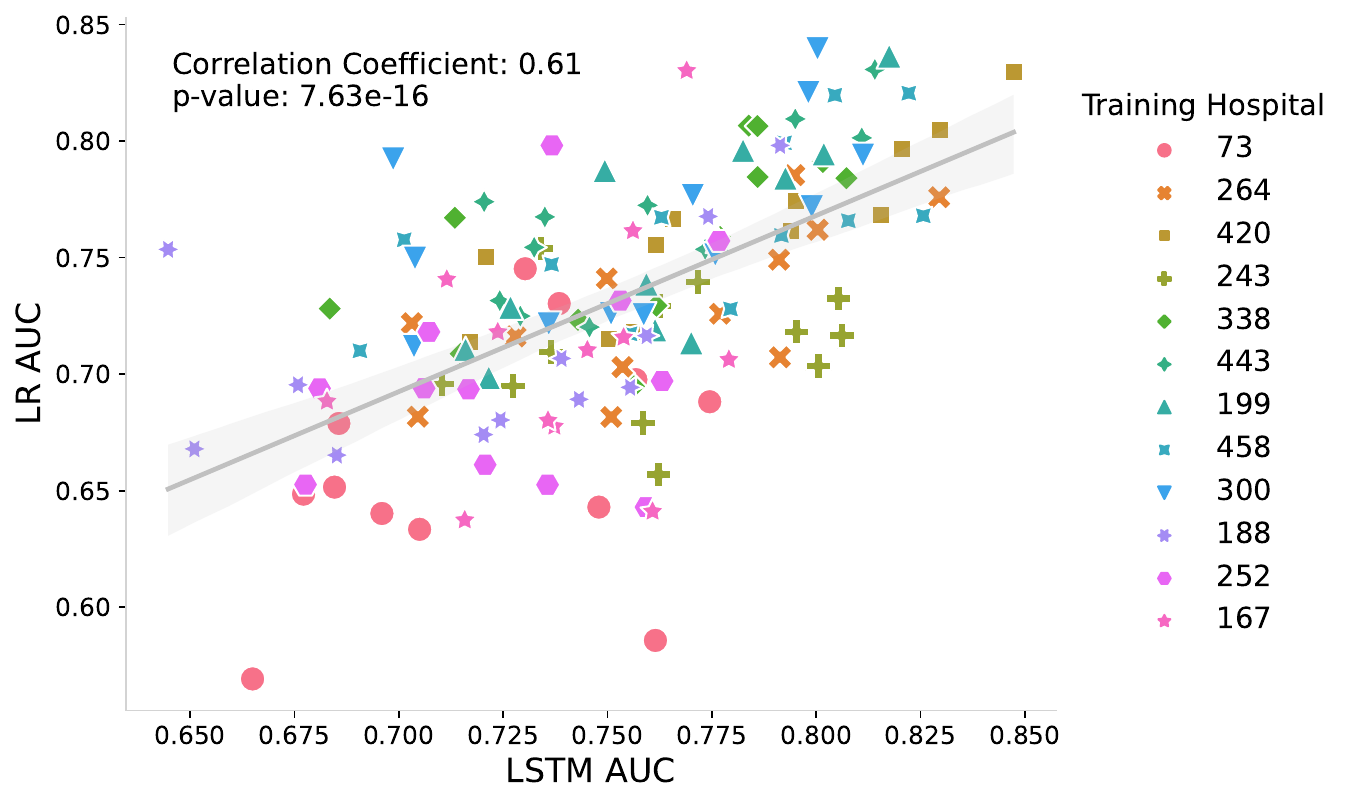}}
    \caption{(a) When training a Logistic Regression model on one hospital ($y$-axis) and testing on another hospital ($x$-axis), the difference in AUC for in-distribution versus out-of-distribution training hospital data can vary across hospital pairs. 
    (b) Changes in AUC due to distribution shifts between training and test (reference) hospital data are generally agnostic to model selection, e.g., the strong and statistically significant correlation of AUC changes between Logistic Regression (LR) and Long Short-Term Memory (LSTM).}
    \label{fig:auc_diff}
\end{figure}

We find that training on one hospital and testing on another yields notable changes in AUC performance (Figure~\ref{fig:auc_diff})---at times making performance slightly better than in-distribution training but more often making performance significantly worse. This complements prior findings that out-of-distribution test sets may yield both positive and negative changes in performance ~\citep{spathis2022looking}. For example, for hospital 73, training on hospital 199 data (instead of hospital 73 data) could yield an improvement of 0.06 AUC; however, for hospital 338, training on hospital 199 data (instead of hospital 338) would yield a drop of 0.04 AUC. These pairwise train-test hospital combinations are consistent across LR, LGBM, and LSTM model (Figure~\ref{fig:auc_diff}b,~\ref{fig:scatter_lr_lgbm}) with strongly significant correlation (LGBM vs LSTM: $0.47, p=\num{2e-9}$; LGBM vs LR: $0.76, p=\num{8e-29}$).
This motivates our data-centric investigation since these models arising from very different training algorithms still yield similar behavior across shifts in distribution between hospitals. 




\subsection{Characterizing Performance Patterns under Data Addition}
\label{sec:addition}
\begin{figure}[ht]
    \centering
\subfigure[]{\label{fig:a}\includegraphics[width=65mm]{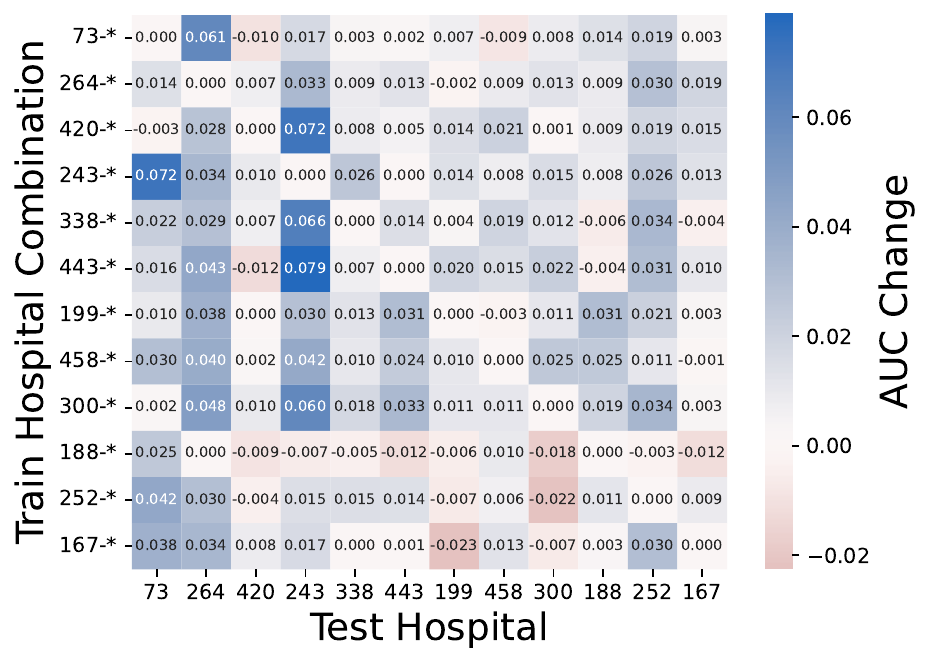}}
\subfigure[]{\label{fig:b}\includegraphics[width=65mm]{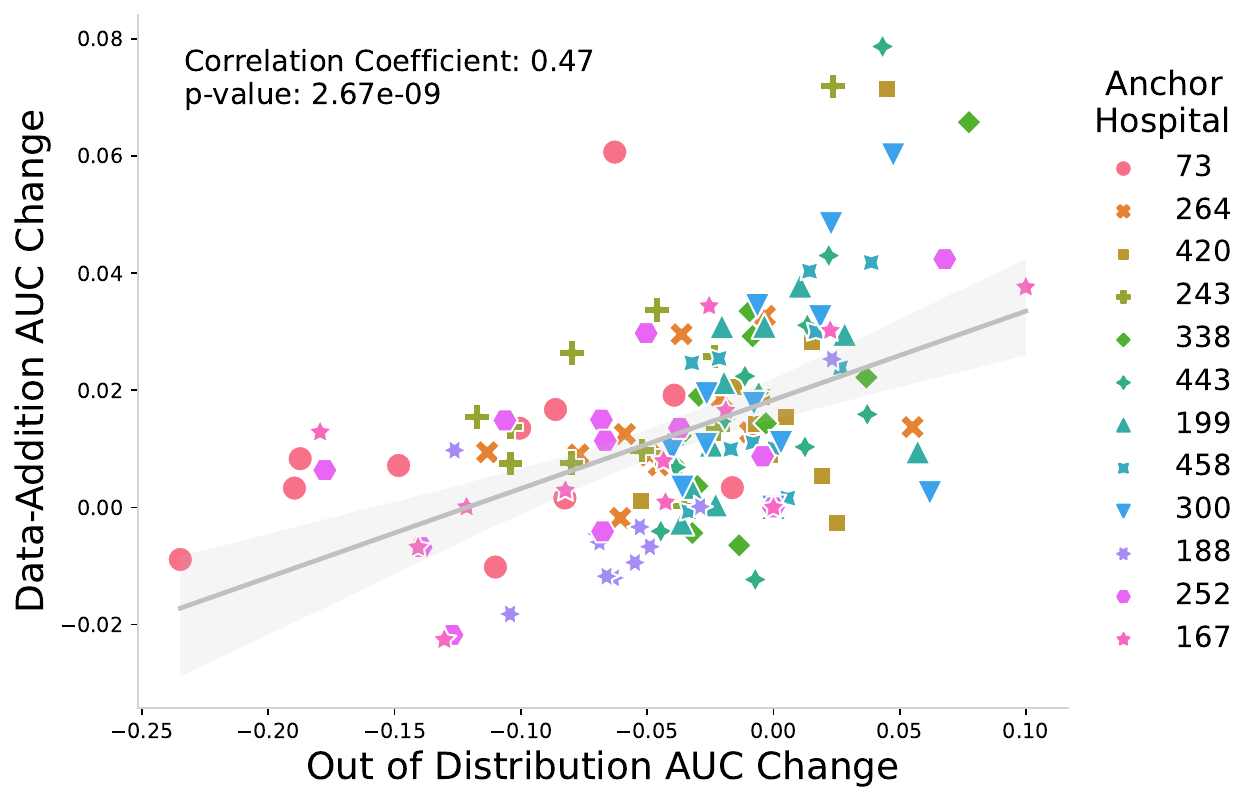}}
    \caption{(a) Training a Logistic Regression model on data from a combination of hospitals can help or hurt AUC on the test hospital. The $y$-axis denotes training on a data combination of a fixed reference hospital (e.g., 73) combined with each of the hospitals in the $x$-axis (e.g., 73 and 264, 73 and 420, etc.) and evaluated on the test data from the reference hospital. The AUC drop is computed by comparing the AUC of training and testing on data from the same hospital (e.g., 73) compared to only training and testing on the test hospital. (b) We find a statistically significant correlation between the data addition AUC and out-of-distribution AUC for training and testing on all pairs of hospitals of anchor hospital and combination hospital using a Logistic Regression model.}
    \label{fig:addition_lr_auc_change}
\end{figure}

Deciding whether to add data sources is a key action for many stakeholders and a primary motivation for our project (Figure~\ref{fig:illustrative}). In order to explore the impact of data addition on data composition changes, we map out the performance pattern changes under a training set composed of the original hospital and an additional source. We find that adding one hospital to the training set yields a variety of AUC changes (Figure~\ref{fig:addition_lr_auc_change}a,~\ref{fig:auc_change_lstm_lgnm}). For example, in the logistic regression model, adding data from hospital 443 can increase the AUC performance on hospital 243 by 0.08; however, adding data from hospital 443 to hospital 420 can instead decrease the AUC performance on hospital 420 by 0.01 (Figure~\ref{fig:addition_lr_auc_change}a). Even more interesting is our finding that the effect of data addition across hospitals remains significantly correlated with the single-source out-of-distribution AUC patterns (0.47, $p < \num{2.7e-9}$)(Figure~\ref{fig:addition_lr_auc_change}b). This gives us signal that observing the performance drops due to distribution shift could be useful in guiding multi-source data addition and scaling. 



\begin{figure}[ht]
    \centering
    \subfigure[]{
        \label{fig:lr_scorexy1}{\includegraphics[width=50mm]{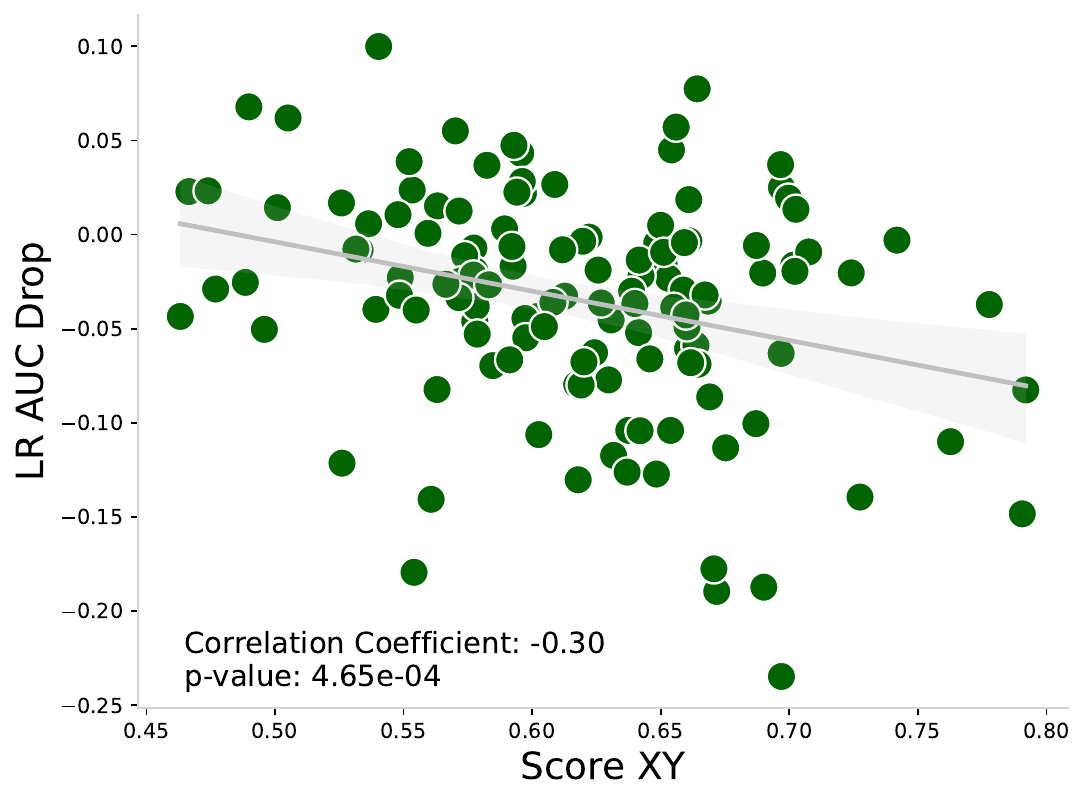}}}
\subfigure[]{\label{fig:all_corr1}\includegraphics[width=90mm]{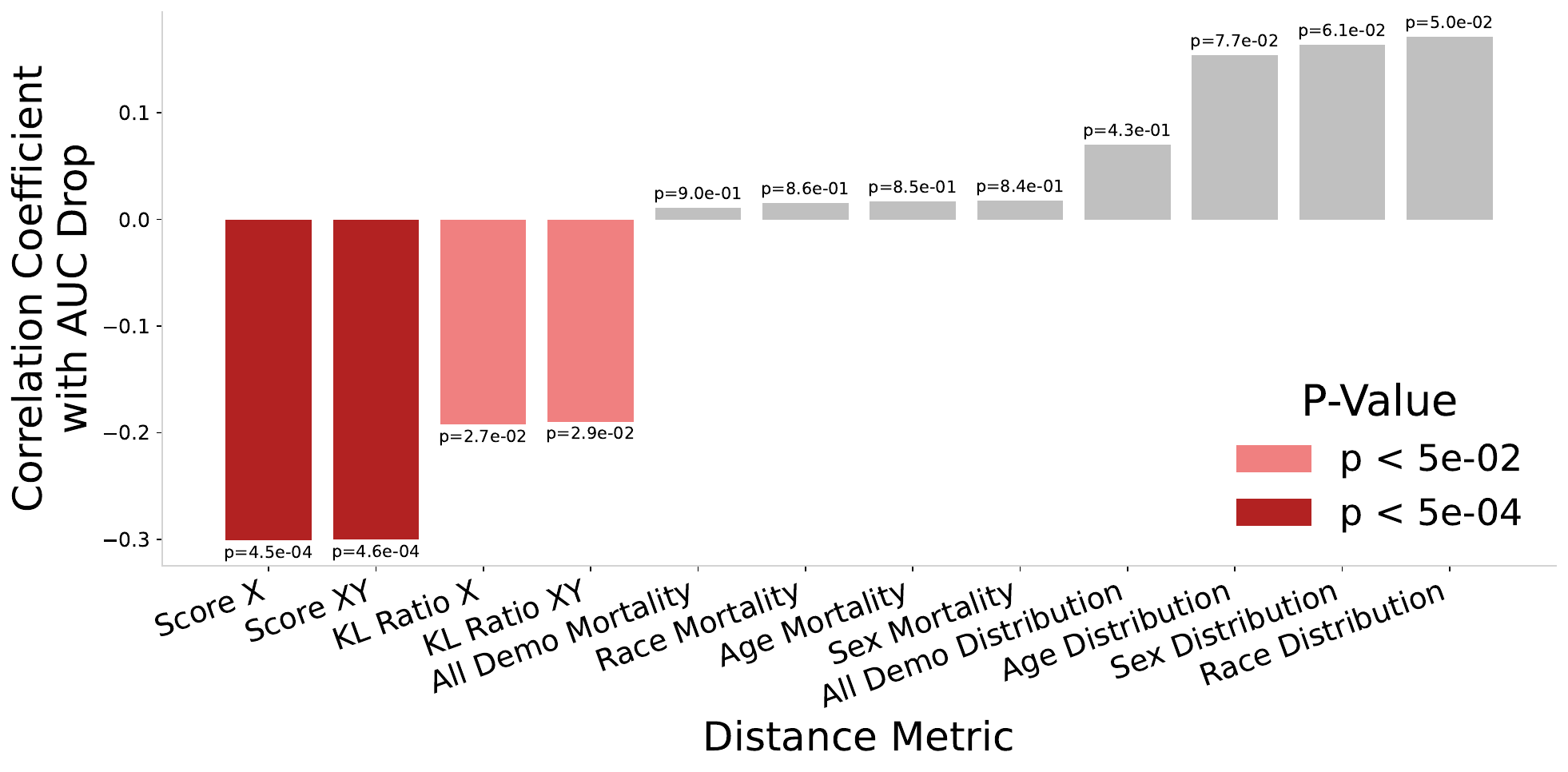}}
    \caption{(a) We find a statistically significant linear correlation between the \textsc{Score XY} metric and the drop in Logstic Regression AUC when training and testing for all pairs of hospitals. (b) We compute the linear correlation between several distance metrics (including \textsc{Score XY}) and the drop in Logistic Regression AUC when training and testing for all pairs of hospitals.}
    \label{fig:dist_metrics}
\end{figure}

\begin{figure}[ht]
    \centering
        \subfigure[]{
        \label{}{\includegraphics[width=65mm]{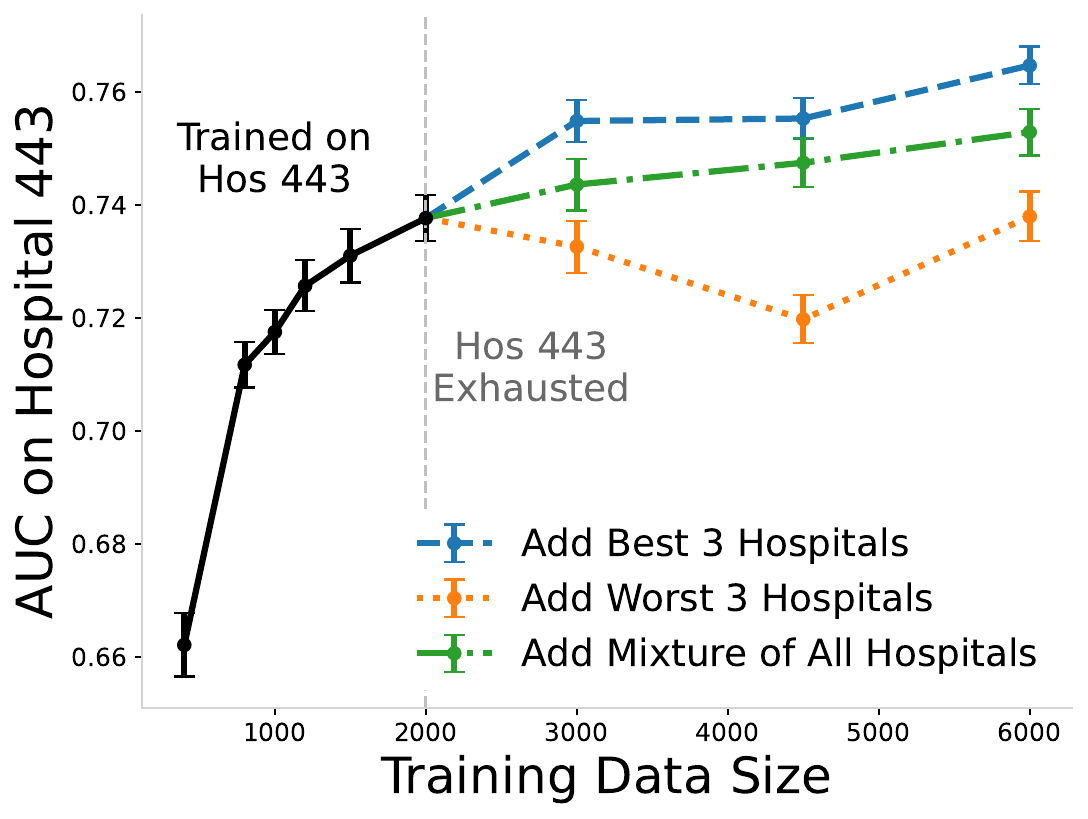}}}
\subfigure[]{\label{}\includegraphics[width=65mm]{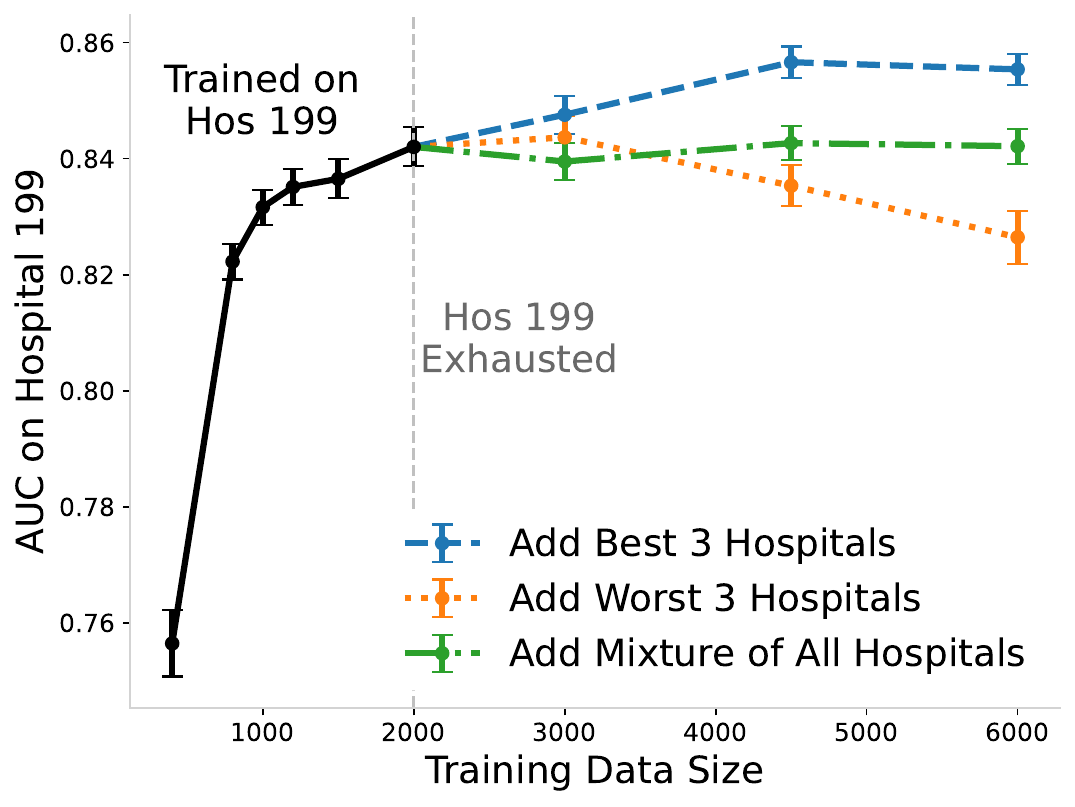}}
    \caption{We compare sequential data addition strategies for (a) hospital 443 and (b) hospital 119 using a Logistic Regression Model. Best 3 and worst 3 hospitals are determined based on the \textsc{Score XY} metric described in Section~\ref{sec:heuristic}.}
    \label{fig:seq_hos443_lr}
\end{figure}

\begin{figure}[ht]
    \centering
            \subfigure[]{
        \label{}{\includegraphics[width=65mm]{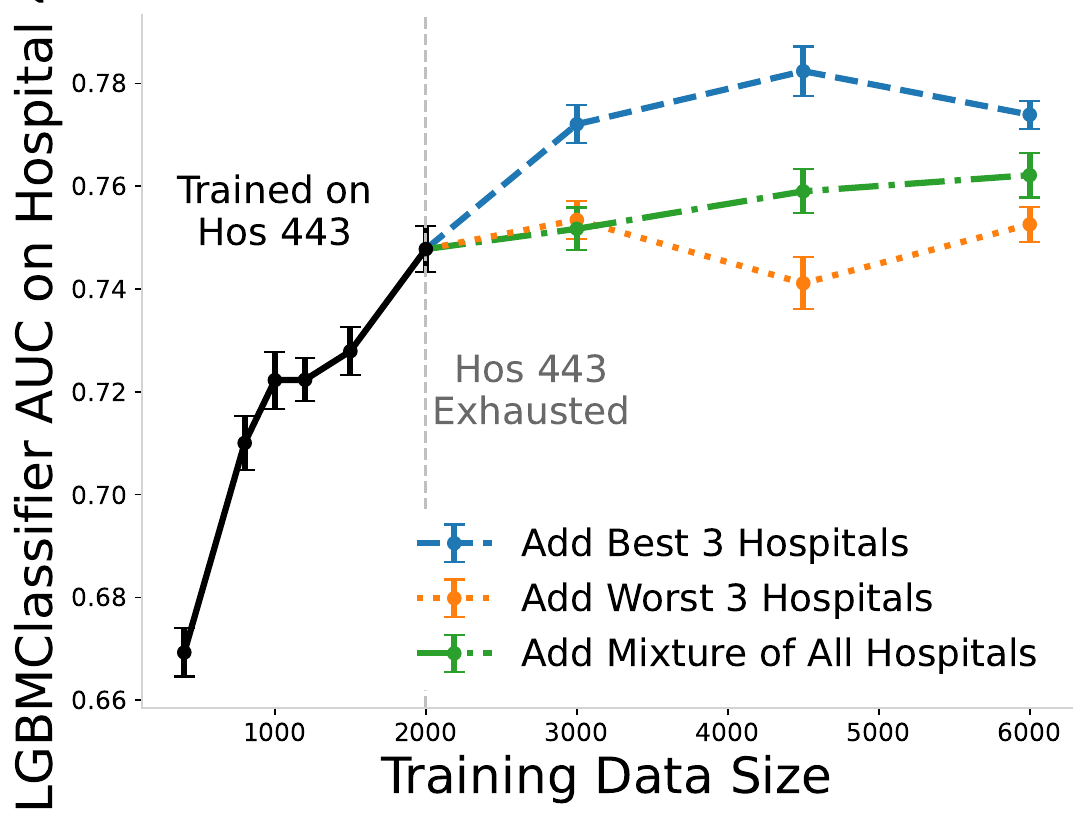}}}
\subfigure[]{\label{}\includegraphics[width=65mm]{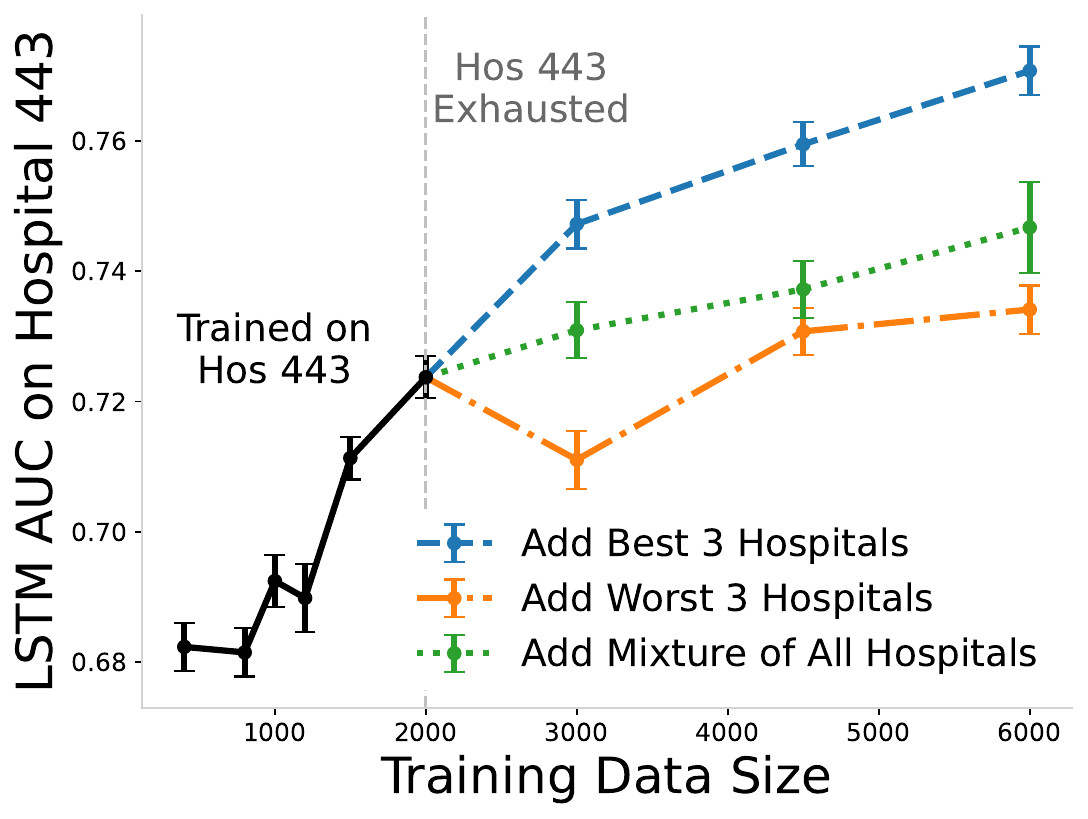}}
    \caption{We evaluate the sequential data addition strategies for hospital 443 using (a) an LGBM Classifier and (b) an LSTM model. Best 3 and worst 3 hospitals are determined based on the \textsc{Score XY} metric described in Section~\ref{sec:heuristic}.}
    \label{fig:seq_hos_lstm}
\end{figure}


\subsection{Data Addition Heuristics}
\label{sec:heuristic}
With the significant correlation between out-of-distribution AUC change and data addition AUC change, we are motivated to find ways to characterize distribution shifts that may help us decide how to add data in the multi-source scaling setting.

We first turn to normative metrics of dataset descriptions that have been studied in prior work for compare distribution shift across and within datasets~\citep{spathis2022looking, schrouff2022diagnosing}. We examine the distance between age, race, and sex distribution, age, race, and sex mortality as well as the overall mortality rate. Firstly, we observe significant differences in these dataset statistics. For example, mortality rates for male patients vary between $1.6\%$ and $9.8\%$ across the 12 hospitals we examine. As another example, the Black patient population varies between $1\%$ and $53\%$. However, it is not known whether these demographic differences are correlated with the severity of out-of-distribution performance (i.e., test time performance drop). Using Euclidian distance between patient distributions and subgroup mortality, we find that there is no significant correlation between the performance drop between train and test hospitals and their distances in summary data statistics, e.g., ``All Demo Mortality'' ($p=\num{9.0e-1}$) or ``Race Distribution'' ($p=\num{5.0e-2}$) (Figure~\ref{fig:dist_metrics}b).

We then evaluate the correlation between our KL and score-based distance metrics with the accuracy drop induced by the distribution shift. Across all 12 hospitals, we find the strongest negative correlation to be between the accuracy drop and our expected score function metric: $(-0.30, p=\num{5e-4})$ (Figure \ref{fig:dist_metrics}a). In other words, as the score and distance between hospital distributions increase between the test and train hospital, the performance on the test hospital decreases. In Figure~\ref{fig:dist_metrics}b, we plot the correlation between all metrics we examine. The gray bars represent normative metrics (dataset statistics distances) from hospital demographics and outcome statistics---none amounted to a significant correlation. We also find that our KL divergence estimates based on density ratio were also negatively correlated with out-of-distribution performance change. Furthermore, although there are differences in outcome distribution across hospitals, we find the correlation between distance metrics computed on just the covariates or features and joint distributions between covariates and targets to be very similar: ($-0.301, p=\num{4.5e-4}$) compared to ($-0.300, p=\num{4.6e-4}$). 

With a compelling distance notion in hand, we can proceed with multi-source scaling using our score metric (\textsc{Score XY}). We study the case of adding three additional hospitals to the training data. There are $\frac{M!}{(M-m)!}$ possibilities for choosing the $m$ additional hospitals from $M$ total hospitals. Instead, we take the most intuitive approach: choosing the closest hospitals according to our distance metric and therefore the most similar hospitals. We also examine adding the furthest hospitals in distance in comparison. As an additional baseline, we sample from a mixture of remaining hospitals with the same training dataset size. 

Figures \ref{fig:seq_hos443_lr} compares the three strategies of: a) adding the three closest (best) hospitals, b) adding the three furthest (worst) hospitals, and c) adding a sample from a mixture of all $M$ hospitals for hospitals 443 and 199 using the Logistic Regression model. For hospital 443, the best three additional hospitals are hospitals 458, 167, and 300 while the worst three are hospitals 73, 252, and 338. For hospital 199, the best three additional hospitals are hospitals 300, 458, and 167 while the worst three are hospitals 73, 252, and 443.

We see that adding the three closest hospitals outperforms both adding the worst three and a mixture across all hospitals. For hospital 199, using all of the available hospital 199 data yields an AUC of 0.842. Adding the worst three hospitals improves AUC (AUC=0.855) whereas adding the worst three hospitals worsens overall performance (AUC=0.826) even though the training dataset size has been expanded. The baseline using a mixture of hospitals is in the middle (AUC=0.842). For the LSTM and LGBM models across the same hospitals, we similarly observe that adding the three closest (best) hospitals performs well (Figure~\ref{fig:seq_hos_lstm}). However, for the LSTM model, even adding the worst three hospitals slightly improves performance for hospital 443. This result is aligned with the common belief that deep learning models more strongly benefit from the addition of data. 

\label{sec:lab}

\subsection{Additional Observations}


In addition to overall model AUC, there are other model performance measures of interest in this study. 
Notably, past investigations indicate that, in a simple tabular census setting, as training data size scales, even when overall model accuracy improves, group disparities may in fact worsen~\citep{ding2021retiring}. We analyze case studies of the impact of data composition and scaling of the \emph{worst group accuracy} and \emph{performance disparity} between groups (as defined in Section ~\ref{sec:exp-models}). More practically, we compute the lowest accuracy measured across all available subgroups as the worst-case subgroup accuracy, and the difference between the worst-case and best-case subgroup accuracy as the subgroup accuracy disparity.




When considering how our results impact subgroup disparity and worst-case subgroup performance, we find that the models perform in fairly unexpected ways (Figure~\ref{fig:seq_hos_disp},~~\ref{fig:seq_hos_worst}). For instance, for hospital 443, the accuracy disparity \emph{decreases} more (0.329 to 0.050) when adding data from the 3 best hospitals using our heuristic compared to our mixture baseline case (0.329  to 0.414), and \emph{increases} worst group accuracy (0.583 to 0.833). However, for hospital 199, the opposite is true---adding data from the three best hospitals according to our heuristic \emph{increases} disparity (0.198 to 0.448) and \emph{reduces} worst group accuracy (0.750 to 0.500), likely due to small sample sizes.
These results are potentially related to model performance differences across groups being inconsistently impacted by the subgroup composition of training sources (Figure~\ref{fig:hos264_auc_subgroups},~\ref{fig:hos420_auc_subgroups}). 
Additionally, under different data addition combinations, worst subgroup performance can vary wildly (Figure ~\ref{fig:addition_lr_acc_change_race0}). Future work could further investigate subgroup performance pattern changes under data addition.



\section{Discussion}
We present the trade-offs to adding more data in the multi-source setting as the Data Addition Dilemma. We motivate this problem by observing that for the commonly studied mortality prediction task in particular, adding data can both improve and worsen performance. We introduce a sequential data accumulation framework to formalize why adding data could yield worse performance. We conduct a thorough investigation into distribution shift and data addition across 12 hospitals and 3 different model architectures. We examine different distances to characterize distribution shifts that better correlate with out-of-distribution performance than population statistics. Ultimately, we find our heuristics to be informative in deciding which sources of data to add. In general, we expect the trade-off between a smaller high-quality data set and a larger low-quality dataset to be model-dependent. Nevertheless, our results motivate practitioners, particularly in high-stakes applications, to carefully consider the costs and benefits of adding more data.

\paragraph{Limitations \& Future Work}
While our work presents a theoretical basis and rigorous experiments of the Data Addition Dilemma, we acknowledge several limitations of our work for a complete picture of a general law for data accumulation.
First, our experimental setting focuses entirely on binary mortality prediction from time-series data in the intensive care unit (ICU) setting, which enables clean performance metrics to compare across hospitals. To build a more complete picture of the trade-offs of data scaling and distribution shift in health and healthcare, we imagine similar principles will apply in future work across key topics like modeling treatment strategies~\citep{gottesman2019guidelines,miao2024identifying}, risk stratification~\citep{razavian2015population,beaulieu2021machine,chen2020intimate,seyyed2021underdiagnosis,seyyed2020chexclusion}, and patient subtyping~\citep{lawton2018developing, choi2023machine, chen2022clustering}. 
Second, our data accumulation method is simplified to comparing adding data from hospitals sequentially with adding a mixture of data from all available hospitals. A more complex model could include a combination of two or more sophisticated data accumulation strategies. 
Thirdly, our work considers data accumulation as the key intervention for the Data Addition Dilemma. There are many other methods to investigate data-driven performance trade-offs including data pruning~\citep{sorscher2022beyond,hooker2020characterising} and synthetic data~\citep{nikolenko2019synthetic}. Lastly, while we present one model for data accumulation, further theoretical work in data modeling such as robustness guarantees~\citep{hu2024multigroup} and distributional analyses~\citep{chaudhuri2023does, feldman2020does} would vastly improve guarantees for decision-making about datasets. 

Data decision-making is a critical factor in the effective execution of machine learning---yet little is understood of how practical data curation and collection strategies ultimately impact model outcomes. This work is an initial inquiry into what is often an overlooked aspect of the machine-learning process. In particular, determining the best protocols for adding health data have important implications for clinical settings. We hope this can be a vehicle towards more thorough future modeling and investigations into the practical data decision-making process that underscores much of the model behavior in deployed systems.

\acks{The authors thank Roshni Sahoo for thoughtful discussions and Ben Kuhn, Nicholas Schiefer, and Sylvia H{\"u}rlimann for visualization brainstorming. This work was supported by Apple Machine Learning, a Google Research Scholar grant, the MacArthur Foundation, the Simons Foundation Collaboration on the Theory of Algorithmic Fairness, and the Simons Foundation Investigators Award 689988.}

\bibliography{references}

\newpage

\appendix
\section{Extensions to Notions of Data Quality}

The canonical model of data in machine learning assumes that empirical data are sampled iid from an underlying distribution. There are several consequences of these given conditions, most notably the following assumptions:
\begin{itemize}
    \item Test and training set data distributions are identical.
    \item Training data distribution remains consistent as the dataset size increases. 
\end{itemize}

However, something that is not as often considered is how the training set distribution \emph{changes} as dataset size increases, at times becoming increasingly divergent from a fixed test distribution. In this work, we specifically investigate this phenomenon.


\subsection{Sources of Sampling Bias}
Not all cheap data is equally bad and not all distributions are equally shifted. In this paper, we discuss sampling bias issues incurred via data accumulation. Here, we awknowledge some common sources of sampling bias that may arise when the following types of data are favored in the data sampling process: 
\begin{enumerate}
    \item Easy-to-access data: Machine learning datasets are often amassed using data that is readily available on the internet or even data that is collected without consent. In the healthcare domain, publicly available datasets from a few select hospitals may dominate the training set for a specific task. 
    \item Complete data: Parts of datasets with missing features may be discarded. Individuals may be included in surveys only if a survey is filled out in its entirety. Differential knowledge of family medical history may impact which health records are used to identify and develop certain diagnostic tools. 
    \item Unambiguous data: Data points with multiple or conflicting labels may be discarded. Outlier or ambiguous images or text data may be discarded to bolder the statistical robustness of the resultant classifier. 
\end{enumerate}
\section{Detailed Related Work}

\begin{figure}
   \centering
   \includegraphics[width=0.7\textwidth]{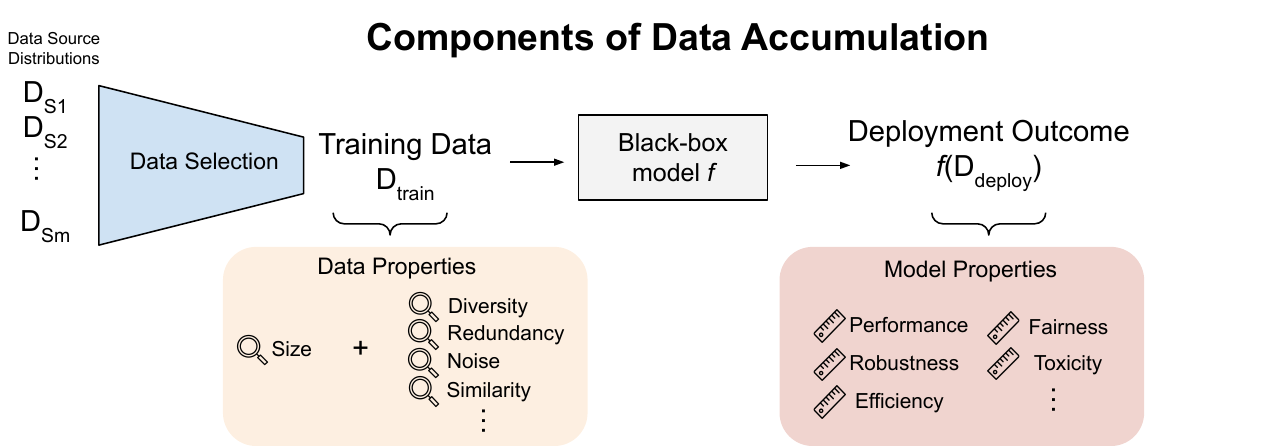}
   \caption{Our work on data accumulation characterizes training data properties interact with scale and how they impacted model outcomes. This study is data-centric and is orthogonal but complimentary to model-based interventions provided in adjacent directions such as domain adaptation.}
   \label{fig:data_accumulation}
\end{figure}

\subsection{Data Scaling}

``Scaling laws'' more broadly refer to how increases in model ``size'' lead to improved performance. Typically, model size is described in terms of the number of model parameters, compute and other measured factors characterizing the model~\citep{kaplan2020scaling,bahri2021explaining}. Data ``scaling laws''~\citep{zhang2020you,bansal2022data,zhai2021scaling} specifically reveal the way in which training on larger and larger datasets yield improved performance. 
For instance, in \citep{touvron2023llama}, large language models trained on datasets with at least 1T tokens beats out models with an order of magnitude more parameters.

Looking at scaling laws for multiple sources in particular, \citep{hashimoto2021model} model data as coming from k sources $q_1, ..., q_k$ where $p=\sum_{k\in[k]}q_kp_k$ is the training data; these sources could be different categories of amazon reviews. They measure the resulting excess loss on some test distribution that arises from training with a specific data mixture of $n$ datapoints $p_{n, q}$: \[
L(n, q) = \Err{l(\hat{\theta}(p_{n, q}, x, y))}{} - \inf_{\theta}\Err{l(\theta; x, y)}{}
\] 
In their experiments, they find that $L(n, q)$ only decreases as $n$ increases; even when $q$ does not match the test distribution well. But they propose an estimation technique based on optimal experimental design which estimates the excess loss based on dataset size $n$ and composition $q$. 

\subsection{The Limitations of More Data}
Unfortunately, more data can also lead us astray. \citep{meng2018statistical} present a model which decomposes estimation error into three components: data quality, data quantity, and problem difficulty: 
\[
\hat{\theta} - \theta = \rho_{R, \theta} \times \sqrt{\frac{1 - f}{f}} \times \sigma_\theta
\]
where $R$ is a binary random variable representing whether an individual responded, $\theta$ is the quantity of interest we are trying to estimate, and $f$ captures how much of the underlying population (i.e. $f=1$ corresponds to the entire population while $f=0$ corresponds to no data). $\rho_{R, \theta}$ represents data quality; if $R$ corresponded to a perfected random sample, this correlation between $\theta$ and $R$ should be zero. $\sqrt{\frac{1 - f}{f}}$ represents error arising from data quantity; if we survey the entire population $f=1$, the error would be zero. $\sigma_{\theta}$  captures problem difficulty and would be zero if $\theta$ is a constant. Furthermore, \citep{meng2018statistical} shows that data quality cannot be compensated by more data when sampling is biased and not truly probabilistic, estimation error scales according to $\sqrt{N}$ the population size and not the sample size. Looking at a $Z-$score for the estimation of $\theta$: 
\[
Z_{n, N} = \frac{\bar{\theta}_n - \bar{\theta}_N}{\sqrt{V_{SRS}(\bar{\theta}_n)}} = \sqrt{N-1} \rho_{R,\theta}
\]
Here we see that while the Z-score should go to zero under random sampling ($\rho_{R,\theta} \approx 0$), the sampling error can scale according to the population size $\sqrt{N}$ when there is sampling bias. 
If we rewrite the above equations with respect to the effective sample size by setting the mean squared error of SRS (simple random sampling) estimator equal to the mean squared error of our biased sampling: 
\[
n_{eff} \leq \frac{f}{1-f} \frac{1}{\Err{\rho_{R, \theta}^2}{R}} = \frac{n}{1-f} \frac{1}{N\Err{\rho_{R, \theta}^2}{R}}
\]
When $\Err{\rho_{R, \theta}^2}{R}$ decreases at a rate of $O(1)$, the effective sample size $n_{eff}$ decreases rapidly. In other words, the actual effective sample size depends crucially on data quality. 

\subsection{Representation matters (Rolf et al, 2021)}

Does collecting more data from underrepresented groups help? In Rolf et al, the key mechanism of change is the allocation of the proportion of groups in the dataset, which the model practitioner can either choose directly or learn for the optimal loss. This assumes the ability to sample directly from the group-specific distribution. The work also maps the approach to importance weighting (reweight training samples with respect to group distributions) and group distributional robust optimization (minimizes the maximum empirical risk over all groups). 

We can formulate this as a function of the group proportions of the training data. For dataset $S = \{ x_i, y_i, g_i \}_{i=1}^n$ where features $x_i$, label $y_i$, and discrete group $g_i$ for groups $G - \{1, \ldots, |G|\}$ are measured. The population prevalence $\gamma_g = P_{(X,Y,G)\sim D} [G=g]$ is related to the ability to empirical allocation of groups in the data $\alpha_g = \frac{1}{n} \sum_{i=1}^n \mathbb{I} [g_i = g]$. 

Sampling from allocation $\alpha$ is defined as independently sampling of $|G|$ disjoint datasets $S_g$ and concatenating according to $S(\alpha,n) = \bigcup_{g \in G} S_g$ with

$$S_g = \{x_i, y_i, g\}_{i=1}^{n_g}, \quad (x_i, y_i) \sim_{iid} D_g$$

Other notions of representativeness such as lexicographic versions of fairness have also been proposed~\cite{henzinger2022leximax}. 
 \subsection{More data can be helpful for fairness sometimes (Chen et al, 2018)}

The paper focuses on many different ways to improve fairness of a model, one of which may be adding more training data. As relevant to this group, error due to variance (as opposed to statistical bias and statistical noise) can be estimated via a distribution learning curve. An assumption of the model is therefore that any new data will be from the same distribution as the training data, on which the learning curve is estimated.

Unfairness can be defined as $\Gamma (\hat{Y}, n) := |\gamma_0(\hat{Y}, n) - \gamma_1(\hat{Y}, n)|$ for predictions $\hat{Y}$, sample size $n$, and group-specific unfairness $\gamma$. Based on prior empirical studies, these type II learning curves can be approximated as asymptotic inverse power-law $\gamma_a(\hat{Y}, n_a) = \alpha_a n_a ^{-\beta_a} + \delta_a$.

However, when subgroups are difficult to identify, the role of data is less known. Recent work by \citet{izzo2023data} presents data-driven strategies for finding subpopulations based on groups where a linear relationship exists between features and the label. While this approach does not explicitly add more data, the composition of data here is imperative for understanding downstream outcomes. 

Given an unbalanced dataset, subsampling (reducing the dataset size as a result) has been shown to achieve better worst-group performance than empirical risk minimization on the entire dataset~\citep{arjovsky2022throwing}. 
\subsection{Relationship to Domain Adaptation, Distribution Shift, and Transfer Learning}

\begin{figure}
    \centering
    \includegraphics[width=\textwidth]{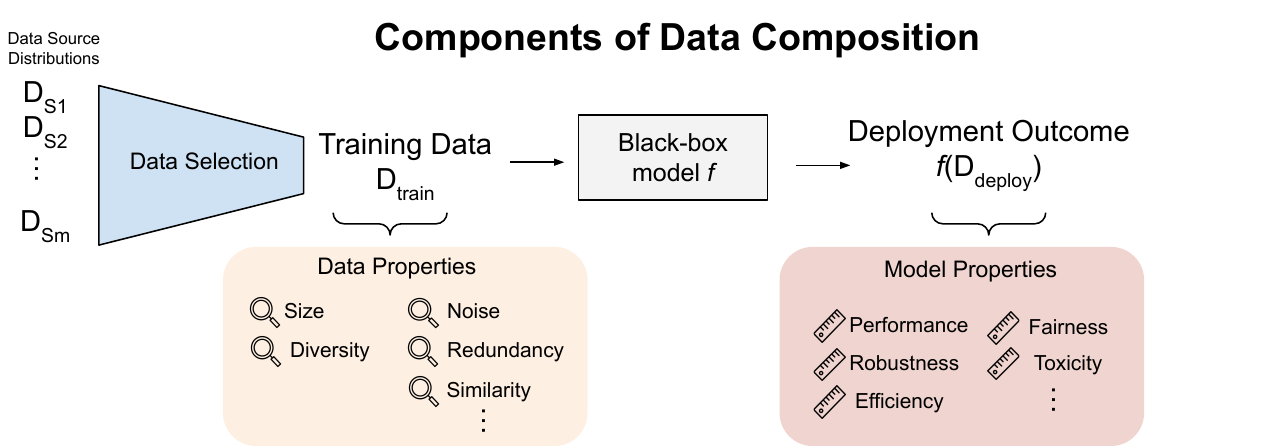}
    \caption{Overview of components of data composition. The domain adaptation literature considers a fixed $D_{train} = \{D_{Source}, D_{Target}\}$ and develops model-based techniques to achieve good \emph{performance} on $D_{deploy} = D_{Target}$ regardless of data properties \citep{kouw2018introduction}. Works in the area of distribution shift assume some difference in \emph{similarity} between $D_{train}$ and $D_{deploy}$ measure drop in \emph{performance} and worst-group performance (a subset of \emph{fairness} metrics \citep{koh2021wilds}.}
    \label{fig:datacomp}
\end{figure}

\paragraph{Domain Adaptation} 
The field of domain adaption provides another lens to view the problem of overcoming data quality challenges. Domain Adaption is concerned with the problem of training models with source data and performing well on a target domain~\citep{kouw2018introduction, zhuang2020comprehensive}. In Unsupervised Domain Adaptation, the majority of the work in the area, labeled source domain data, and unlabeled target domain data are available for training\citep{wilson2020survey}. In Supervised Domain Adaptation, only a few or scarce samples from the target dataset and the source dataset are fully labeled\citep{motiian2017unified}. In this setting, the target domain can be thought of as high-quality data, and the source domain can be thought of as low-quality data~\citep{kouw2018introduction}.

A related area of literature concerned with shifts in data quality is distributional robustness and distribution shift. Worst subgroup performance (subpopulation shift) and overall performance (domain generalization) on an out-of-distribution dataset are metrics presented in the domain adaptation literature aimed at measuring distribution shift~\citep{koh2021wilds}. Methods for domain generalization have been critiqued as no better than empirical risk minimization ~\citep{gulrajani2020search}. Spurious correlation and unseen data shift observed by ~\citet{wiles2021fine} can be thought of as a fine-grained analysis of barriers for domain generalization. Subpopulation shift, the type of distribution shift, describes a difference in subgroup prevalence between training and unseen datasets that might result in compromised worst group performance~\citep{yang2023change}. Works in this area consider a fixed dataset for training(e.g. In- \& Out-of-Distribution) to develop algorithmic interventions and give theoretical guarantees based on a uniform sample size or asymptotic results. Little is known about how sample size and data composition contribute to distributional differences themselves.   

\paragraph{Data Composition} A recent line of work has looked beyond shifts in train and test distributions and instead asked the question of how properties of datasets such as size, diversity, noise, redundancy, and similarity affect model outcomes. \emph{Data Composition} is a broader framework for how data impacts outcomes rather than algorithmic interventions to achieve certain model outcomes (Figure \ref{fig:datacomp}). For example, \citet{nguyen2022quality} study vision-language models and how combining datasets and increasing dataset size impacts overall performance and distributional robustness. In fact, \citet{gadre2023datacomp} introduces a common task where given a fixed set of models, the goal is to find the best subset of training data. Table \ref{tab:data_comp} summarizes existing work in data composition in terms of dataset domain, data properties measured and model outcomes of interest.

\begin{table}[]
    \centering
    \begin{tabular}{l l p{1.2in} p{1.7in}}
    \toprule
     &  Domain &  Data Properties & Model Outcomes\\ \midrule
     \citet{nguyen2022quality}    & Vision-Language  & Noise, Size & Robustness \\ 
     \citet{xie2023doremi}    & Language  & Diversity & Performance, Efficiency \\ 
     \citet{marion2023less} & Language & Redundancy & Performance \\ 
     \citet{gadre2023datacomp} & Multi-modal & Size, Similarity, Redundancy & Performance \\ 
     \midrule
     Our Work & Tabular Data & Similarity, Size & Performance, Fairness (Disparity), Robustness (Worst Group)\\
     \bottomrule
    \end{tabular}
    \caption{Summary of Existing Work in Data Composition}
    \label{tab:data_comp}
\end{table}

\begin{table*}[t]
    \centering
    \begin{tabular}{p{1.1in}rp{0.9in}p{0.6in}p{0.6in}}
    \toprule
        {\sc Dataset} & {\sc Number of Rows} & {\sc Outcome} & {\sc Source} & {\sc Subgroup} \\ 
        \midrule
        eICU \mbox{\citep{pollard2018eicu}} &  200,859 & Binary \mbox{Readmission} &  Hospital & Race, Sex, and Age \\
\midrule
                MIMIC-IV \mbox{\citep{johnson2020mimic}} &  197,756 & Binary \mbox{Readmission} &  Admission Type & Race \\
\midrule
        Folktables \mbox{\citep{ding2021retiring}} & 1,664,500  & Binary Income Level & State & Race \\
        \midrule
        Yelp \mbox{\citep{yelpds}} &  6,990,280 & Multi-Class \mbox{Review} Stars    & State & Restaurant Category \\
        \midrule
        \bottomrule
    \end{tabular}
    \caption{Dataset overview for experiments.}
    \label{tab:data_overview}
\end{table*}

\section{Additional eICU Experiment Results}

We include additional experiment results on eICU data. 

\begin{figure}
    \centering
    \subfigure[]{
        \label{fig:blank}{\includegraphics[width=65mm]{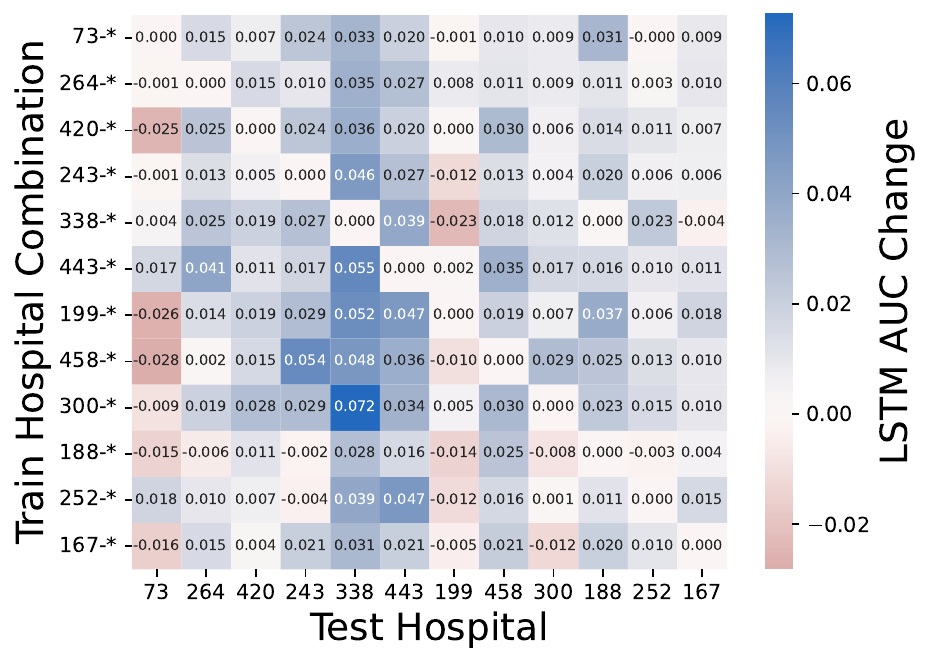}}}
    \subfigure[]{\label{fig:blank}\includegraphics[width=65mm]{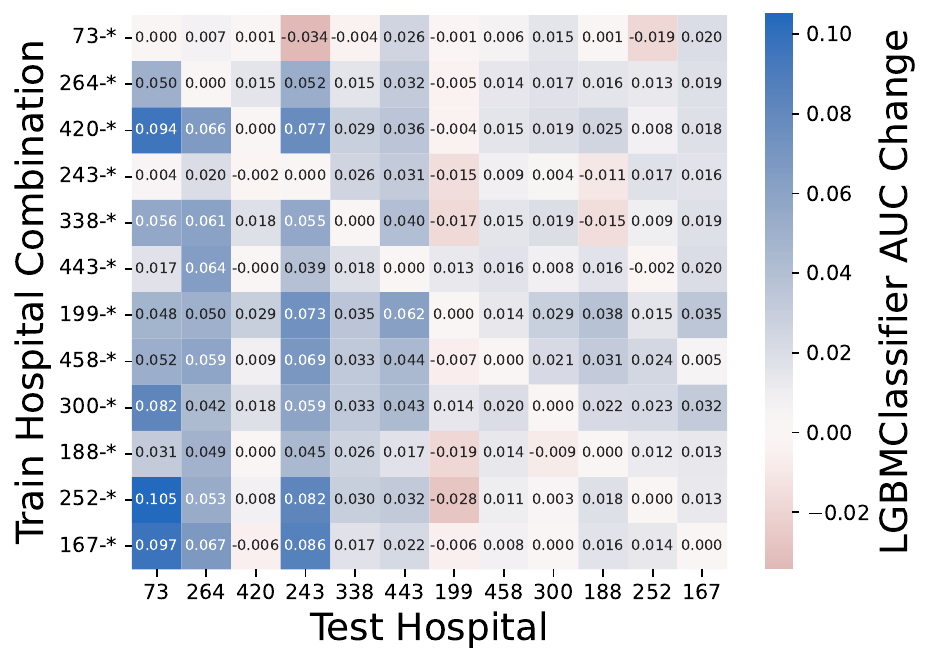}}
    \caption{Changes in AUC performance when adding hospitals for LSTM and LGBMClassifier models. The y-axis denotes training on a data combination of a fixed hospital (e.g., 73) combined with each of the hospitals in the y-axis (e.g., 73 and 264, 73 and 420, etc.)}
    \label{fig:auc_change_lstm_lgnm}
\end{figure}

\begin{figure}
    \centering
    \includegraphics[width=0.4\textwidth]{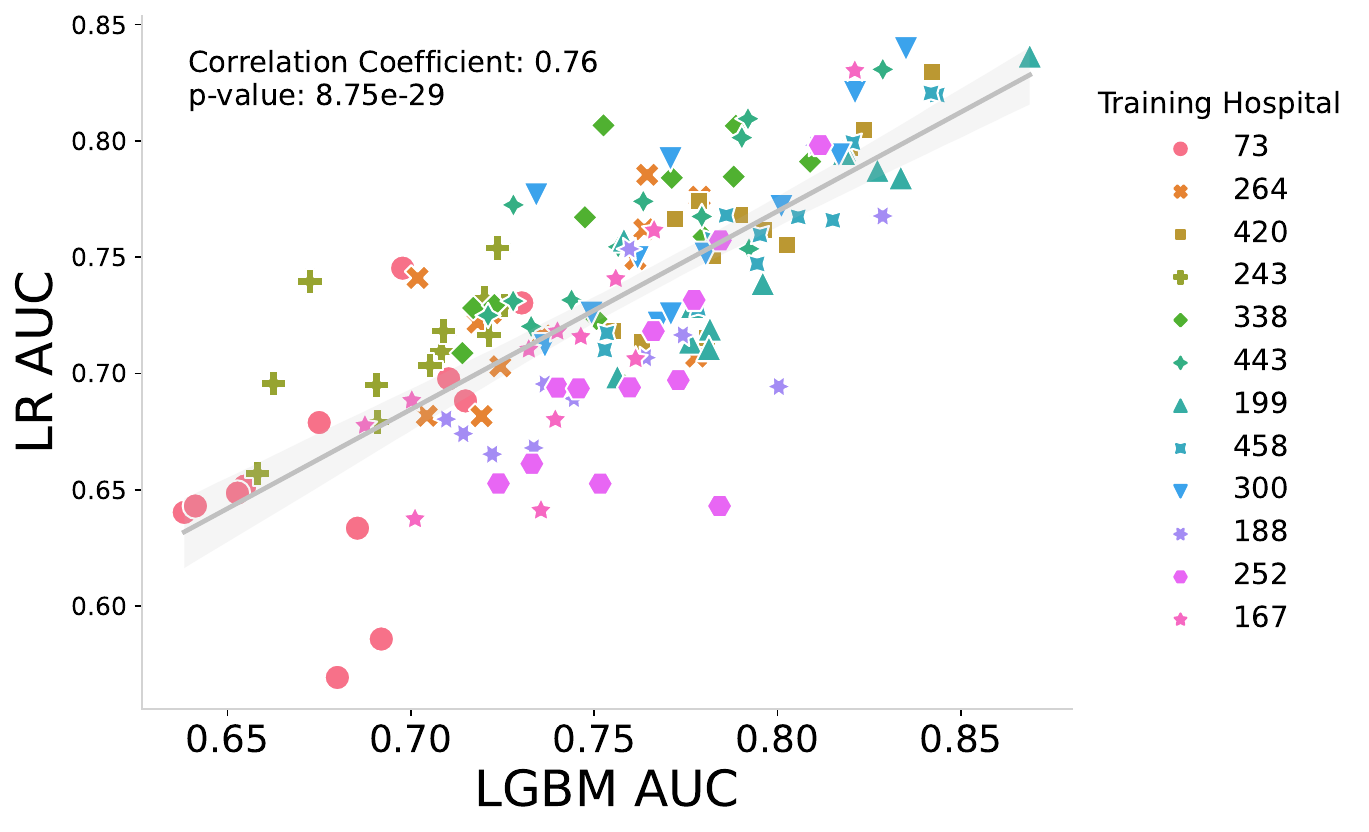}
    \caption{We find a strong correlation between Logistic Regression AUC and LBGM Classifier AUC when training on one hospital and testing on another for all pairwise hospitals. We train models on 1500 samples from each available training hospital and assess the AUC on 400 samples from the test hospital.}
    \label{fig:scatter_lr_lgbm}
\end{figure}

\textbf{AUC Change for additional models} In the main text, we present the results of adding an additional hospital to the training data for logistic regression (Figure ~\ref{fig:addition_lr_auc_change}). In Figure~\ref{fig:auc_change_lstm_lgnm}, we show that both LSTM and LGBM models can induce decreases or increases in AUC when adding combinations of hospitals. In Figure~\ref{fig:scatter_lr_lgbm}, we plot the correlation between the Logistic Regression AUC and LGBM Classifier AUC for the same train-test pair of hospitals. 

\begin{figure}[ht]
    \centering
    \includegraphics[width=0.9\textwidth]{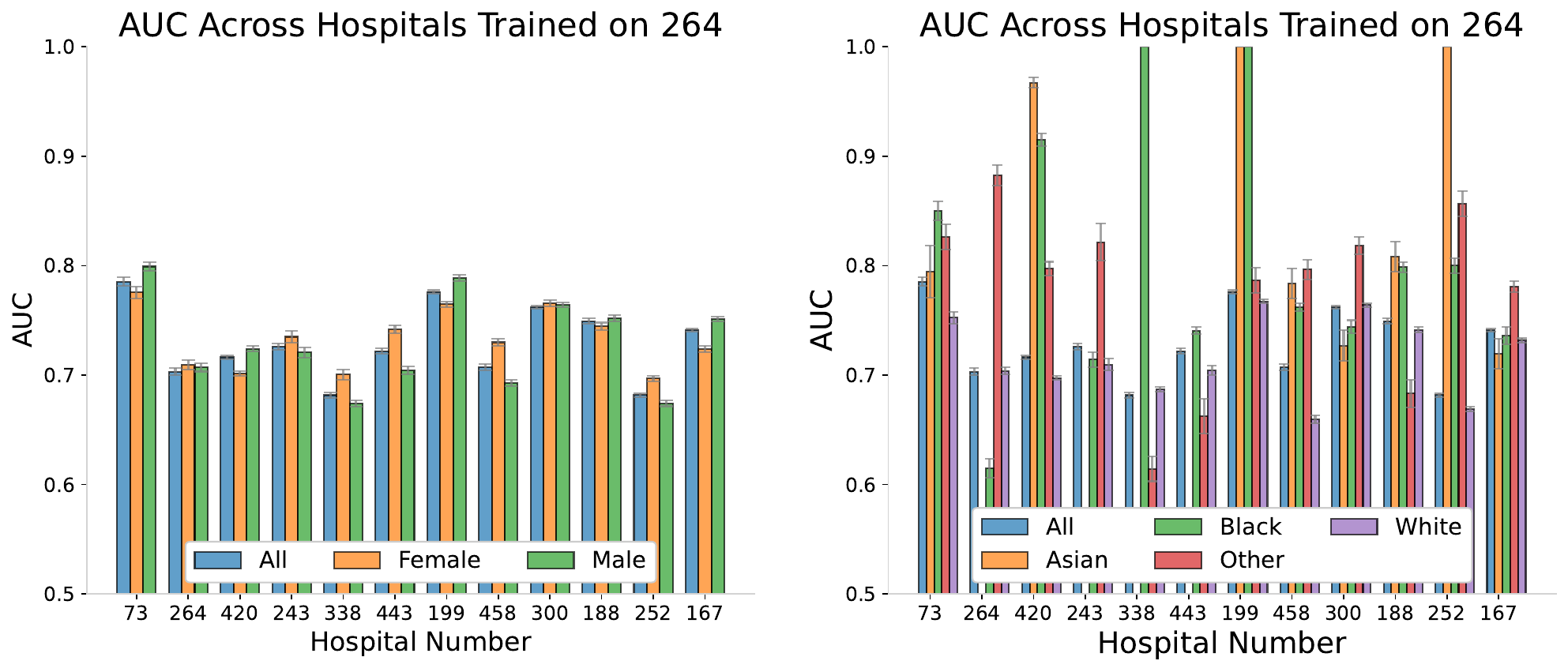}
    \caption{Subgroup Performances for AUC for Hospital 264}
    \label{fig:hos264_auc_subgroups}
\end{figure}

\begin{figure}[ht]
    \centering
    \includegraphics[width=0.9\textwidth]{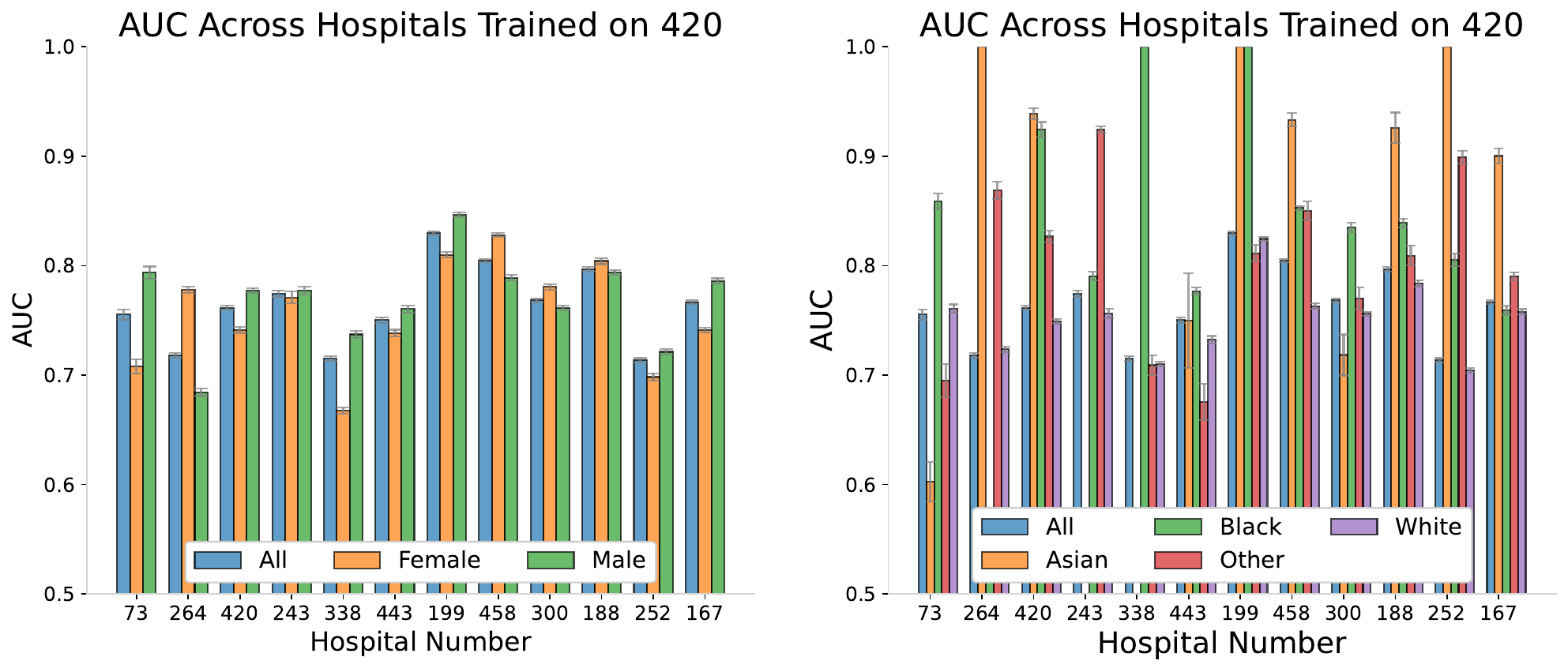}
    \caption{Subgroup Performances for AUC for Hospital 420}
    \label{fig:hos420_auc_subgroups}
\end{figure}

\begin{figure}[ht]
    \centering
    \subfigure[]{\label{}\includegraphics[width=65mm]{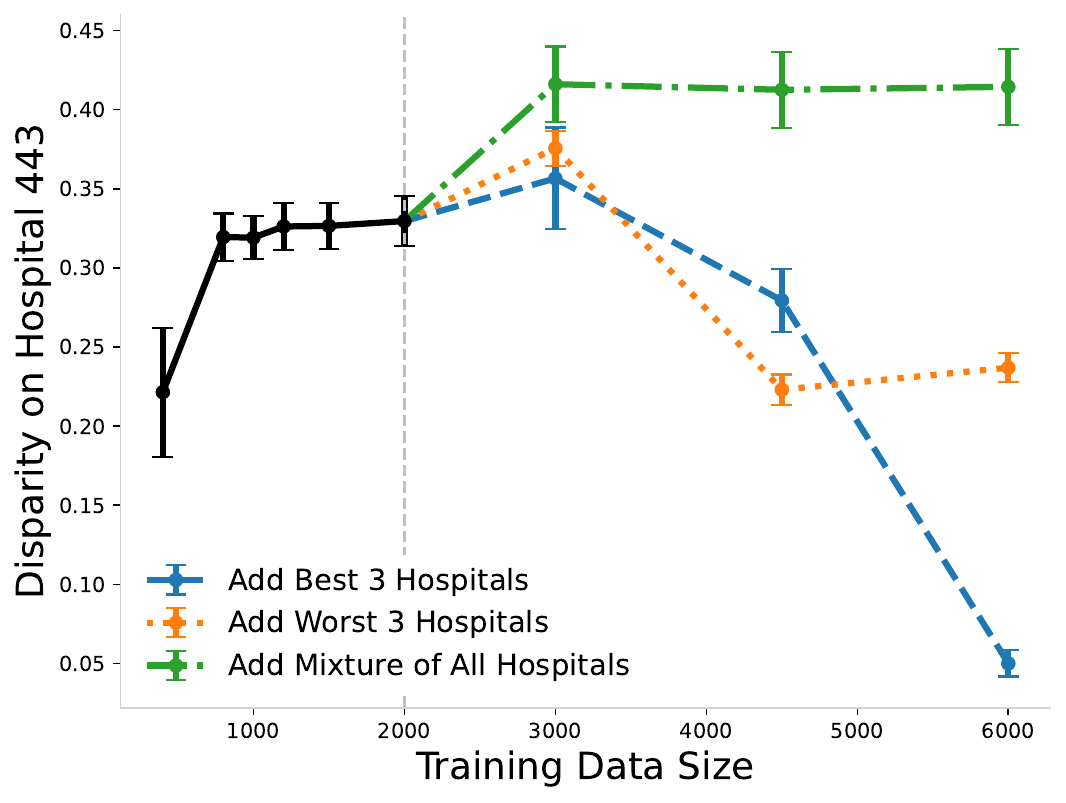}}
\subfigure[]{\label{}\includegraphics[width=65mm]{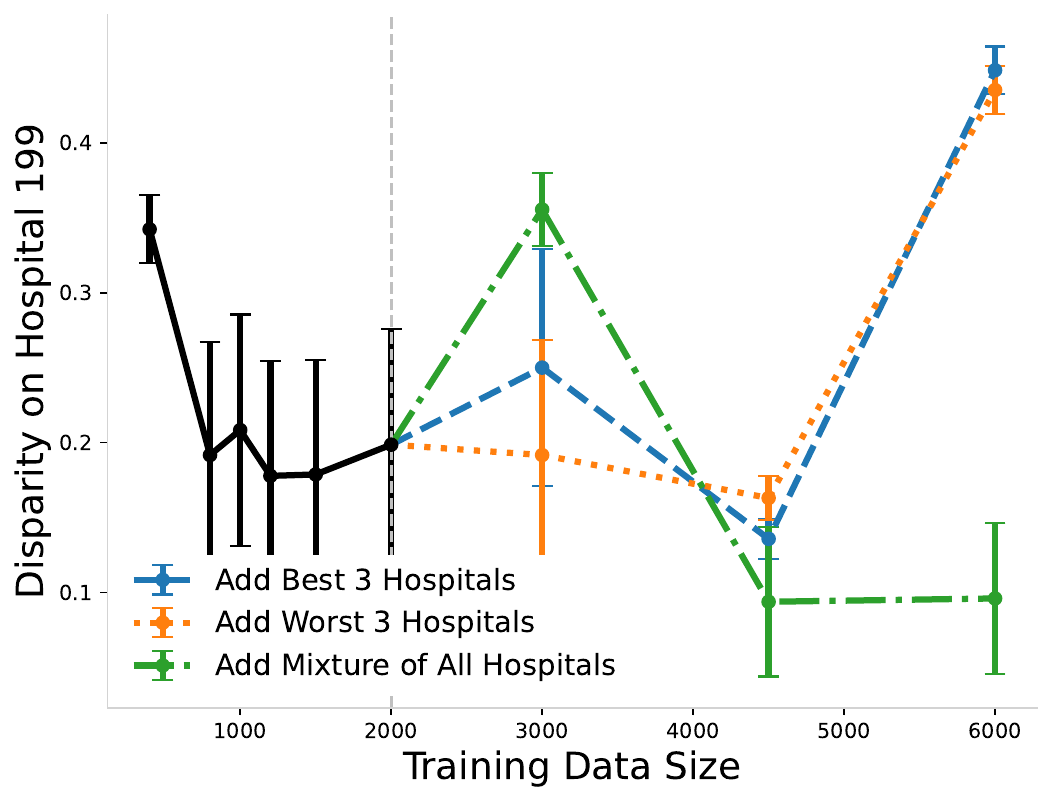}}
    \caption{Effects of Sequential Data Addition Strategies on Race Disparity for Hospitals 443 and 119.}
    \label{fig:seq_hos_disp}
\end{figure}

\begin{figure}[ht]
    \centering
    \includegraphics[width=0.6\textwidth]{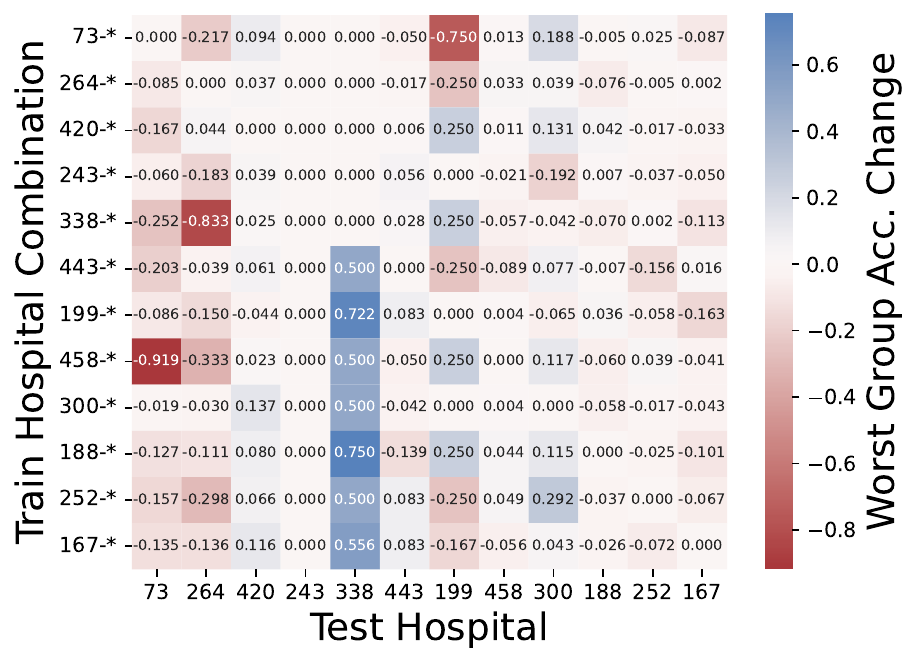}
    \caption{Logistic Regression Acc Drop For Data Addition for Worst Group}
    \label{fig:addition_lr_acc_change_race0}
\end{figure}

\begin{figure}[ht]
    \centering
    \subfigure[]{\label{}\includegraphics[width=65mm]{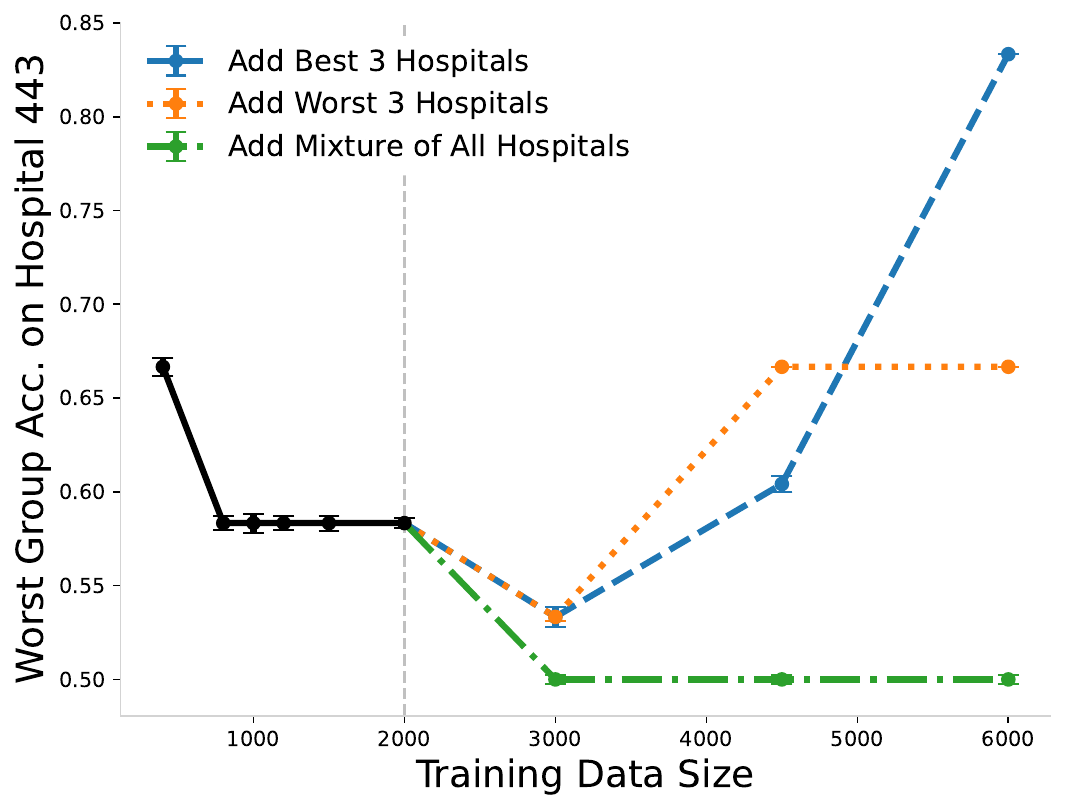}}
\subfigure[]{\label{}\includegraphics[width=65mm]{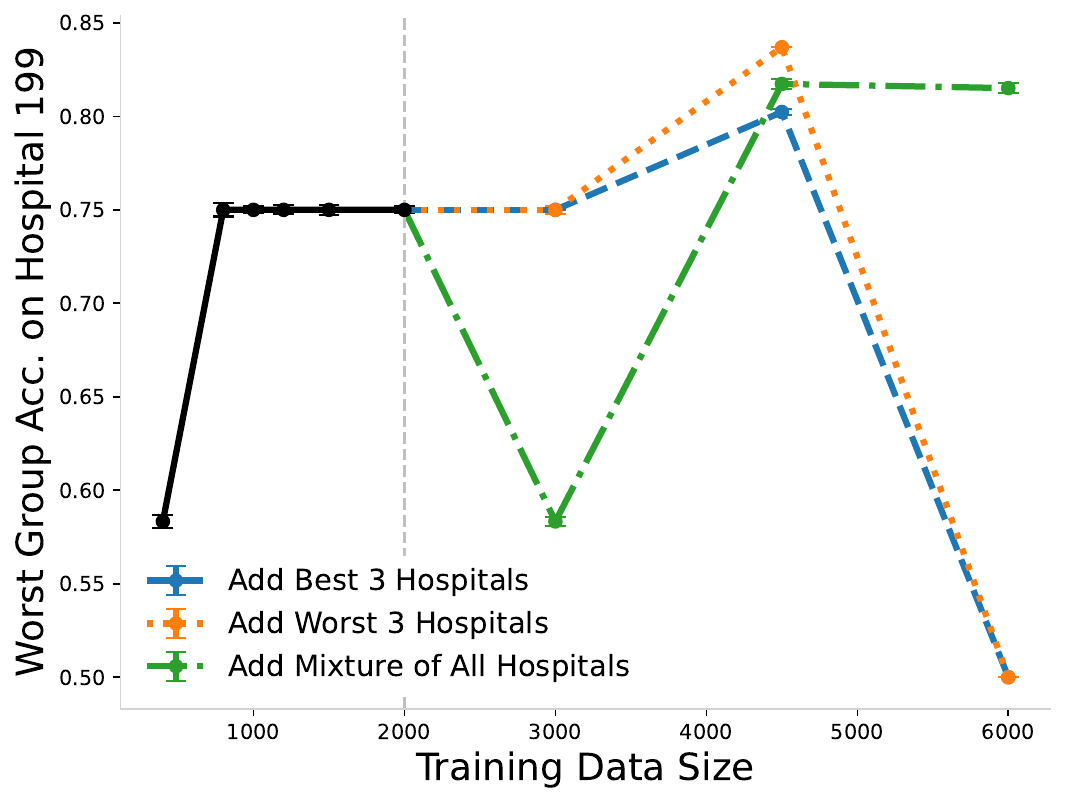}}
    \caption{Effects of Sequential Data Addition Strategies on Worst Race Group Accuracy for Hospitals 443 and 119.}
    \label{fig:seq_hos_worst}
\end{figure}

\textbf{Disparity and Worst Group Performance} We include supplementary plots for disparity and worst group analysis. In Figure~\ref{fig:addition_lr_acc_change_race0}, we describe the impact of combining hospital data on the worst group accuracy for Logistic Regression. In Figure~\ref{fig:seq_hos_disp}, we examine the impact on disparity for hospitals 443 and 199 when adding training data from the three best hospitals as identified in Section~\ref{sec:heuristic}. These results are likely due to differences in AUC across demographic groups (Figure~\ref{fig:hos264_auc_subgroups} and ~\ref{fig:hos420_auc_subgroups}).

\textbf{Smaller Dataset Sizes} Our main paper results demonstrating the effectiveness of data selection heuristics (Figure~\ref{fig:seq_hos443_lr} etc.) cap the sample size of hospitals at $n=1500$, we also conduct ablation studies examining the effect of smaller dataset sizes ($n=1200$, $n=1000$). Figure~\ref{fig:seq_hos_disp} shows that also at smaller data source size, we also observe a significant gap between performance when adding the best and worst hospitals according to our score heuristic. 

\begin{figure}
    \centering
    \includegraphics[width=\linewidth]{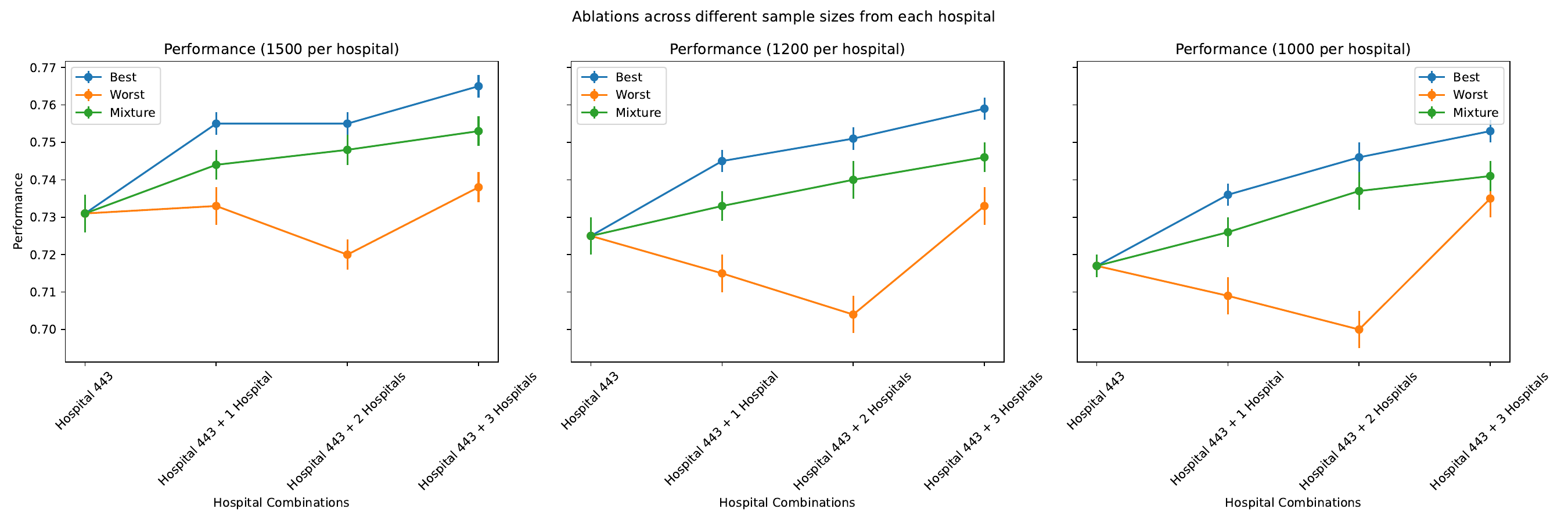}
    \caption{Result of adding best and worst hospitals when dataset size is limited to 1500, 1200, and 1000 samples per hospital for hospital 443. We see that similar behavior occurs when additional data is limited to a smaller size.}
    \label{fig:hos_size_ab}
\end{figure}

\textbf{Consistency of score functions across groups} 
The distances from the scores function we presented in Figure~\ref{fig:dist_metrics} are computed on 2000 samples. Since the motivation of our work is the scenario where accessing other datasets can be costly beyond a small demo dataset, we also test the correlation of distances computed on score functions computed on 2000 samples and smaller sample sizes. Table~\ref{tab:correlation_score} shows that correlation remains fairly high until only 50 samples are used to compute the score function. 

\begin{table}[h!]
    \centering
    \begin{tabular}{|c|c|}
        \hline
        \textbf{Sample Size (n)} & \textbf{Pearson Correlation with 2000 sample ground truth} \\
        \hline
        1500 & 0.994 (p=0) \\
        \hline
        1000 & 0.994 (p=0) \\
        \hline
        800 & 0.993 (p=0) \\
        \hline
        500 & 0.994 (p=0) \\
        \hline
        200 & 0.988 (p=0) \\
        \hline
        100 & 0.972 (p=0) \\
        \hline
        50 & 0.955 (p=0) \\
        \hline
    \end{tabular}
    \caption{Correlation between computing score functions with the full 2000 sample vs smaller data sizes}
    \label{tab:correlation_score}
\end{table}

\begin{figure*}
    \centering
    \includegraphics[width=0.9\textwidth]{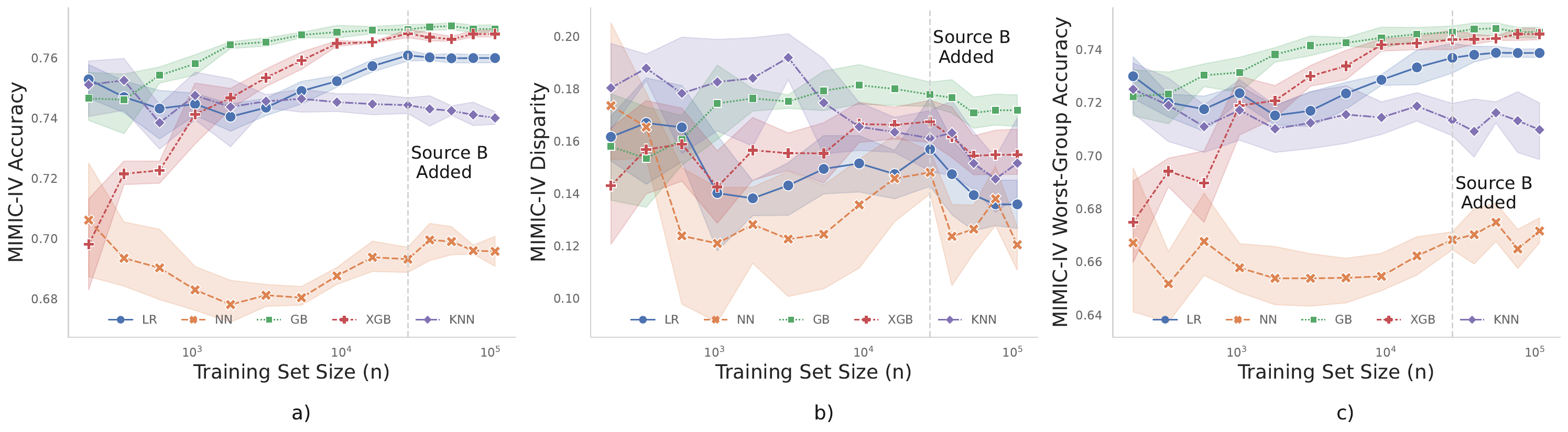}
    \caption{MIMIC-IV results on \textbf{Source A reference test set} in the \textsc{Sequential} case over 5 trials on  \textbf{(a)} accuracy, \textbf{(b)} accuracy disparity, and \textbf{(c)} worst subgroup accuracy. Source A is from admission type URGENT and Source B is from EW. EMER.}
    \label{fig:mimic_dip}
\end{figure*}

\begin{figure*}
    \centering
    \includegraphics[width=0.9\textwidth]{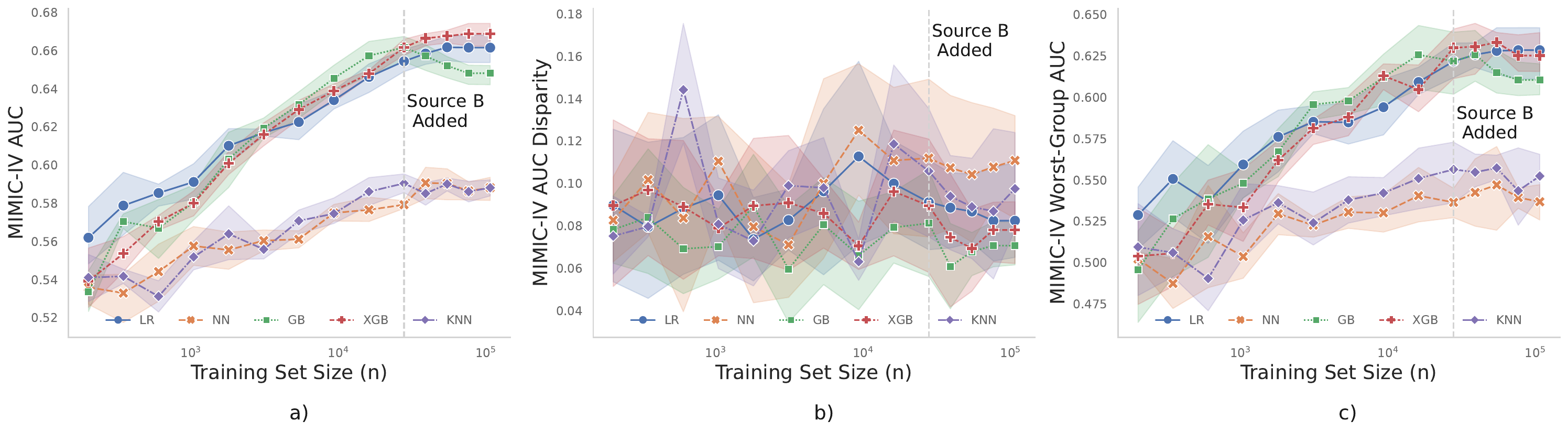}
    \caption{AUC for MIMIC-IV results on \textbf{Source A reference test set} in the \textsc{Sequential} case over 50 trials for \textbf{(a)} changes in AUC with increasing data, \textbf{(b)} changes in AUC disparity across subgroups, and \textbf{(c)} worst case subgroup AUC.}
    \label{fig:mimic_seq_AUC}
\end{figure*}

\section{Additional Dataset Results \& Analysis}


\begin{table*}[t]
    \centering
    \begin{tabular}{p{1.1in}rp{0.9in}p{0.6in}p{0.6in}}
    \toprule
        {\sc Dataset} & {\sc Number of Rows} & {\sc Outcome} & {\sc Source} & {\sc Subgroup} \\ 
        \midrule
        Folktables \mbox{\citep{ding2021retiring}} & 1,664,500  & Binary Income Level & State & Race \\
        \midrule
        Yelp \mbox{\citep{yelpds}} &  6,990,280 & Multi-Class \mbox{Review} Stars    & State & Restaurant Category \\
        \midrule
        MIMIC-IV \mbox{\citep{johnson2020mimic}} &  197,756 & Binary \mbox{Readmission} &  Admission Type & Race \\
        \bottomrule
    \end{tabular}
    \caption{Dataset overview for experiments.}
    \label{tab:data_overview}
\end{table*}

\paragraph{Datasets}
For our investigation, we study three real-world tasks and datasets, chosen because of their rich feature sets and the open-access availability. Dataset details can be found in Table~\ref{tab:data_overview}.

\paragraph{Models and Evaluation}


We consider the scenario where the initial dataset of interest comes from a single data source, i.e., Source A, with limited training examples (e.g., South Dakota (SD) in Folktables). We also consider at least one second available data source, i.e., Source B, from which additional training examples can be drawn. The experimental goal is to investigate the effects of manipulating training data composition on model outcomes as measured on a test set sampled exclusively from the initial dataset Source A  (e.g., SD) – which we call the \emph{reference test set}. Further experiments are also evaluated on a \emph{generalized test set}, which is randomly sampled from a mixture of all available data sources. 

To observe data accumulation in the \textsc{Mixture} case, Source A and Source B are sampled at a fixed ratio from a combined dataset of Source A and Source B in order to increase the training set size. In the \textsc{Sequential} case, we start by adding training data from Source A and then from Source B when all points from Source A have been included. 

We consider a variety of different models: logistic regression (LR), gradient boosting (GB)~\citep{friedman2001greedy}, k-Nearest Neighbors (kNN), XGBoost (XGB)~\citep{Chen:2016:XST:2939672.2939785}, and MLP Neural Networks (NN). Let $f$ denote the model we are evaluating and let $g$ be a group function that maps each data point to a subgroup, we evaluate the following metrics over $D_{test}$:  \textbf{Accuracy}: ($\Er{f(x) = y}{(x, y) \sim D_{test}}$), \textbf{Disparity}: The difference between the best and worst-performing subgroups  ($\max_{g'} \Er{f(x) = y |g(x) = g'}{(x, y) \sim D_{test}} - \min_{g'}\Er{f(x) = y |g(x) = g'}{(x, y) \sim D_{test}}$), and \textbf{Worst group accuracy}: The accuracy on the worst-performing subgroup and the metric of interest in studying subpopulation shifts in the distribution shift literature~\citep{koh2021wilds} ($\min_{g'}\Er{f(x) = y |g(x) = g'}{(x, y) \sim D_{test}}$).

\begin{figure*}
    \centering
    \includegraphics[width=0.9\textwidth]{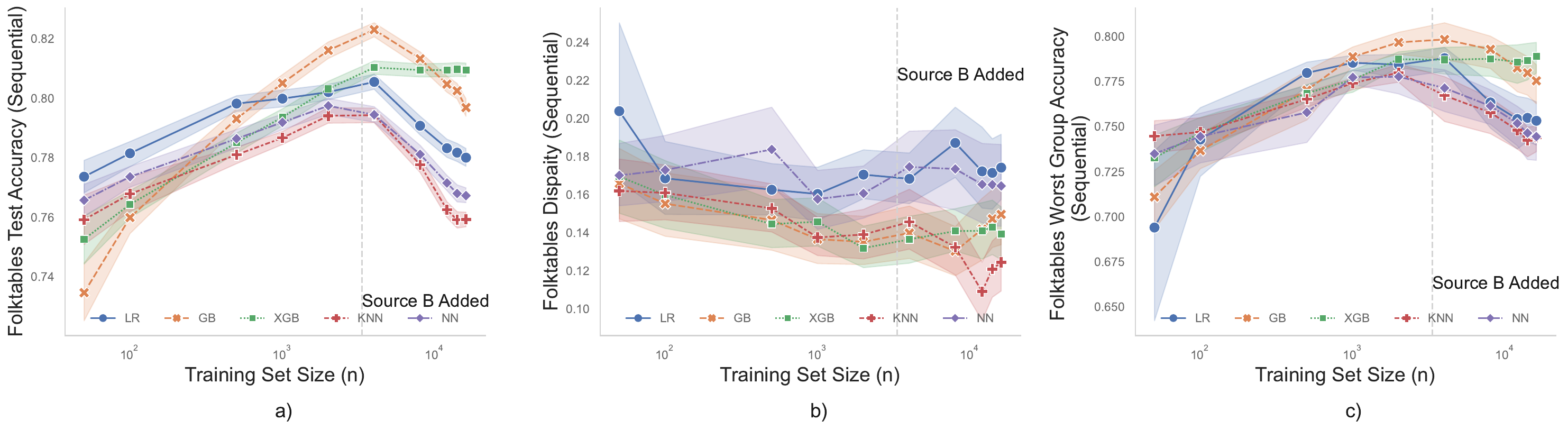}
    \caption{Folktables results on \textbf{Source A reference test set} in the \textsc{Sequential} case over 5 trials on  \textbf{(a)} accuracy, \textbf{(b)}  accuracy disparity, and \textbf{(c)} worst subgroup accuracy. Source A from South Dakota and Source B from California is added once South Dakota data has been exhausted.}
    \label{fig:folktables_seq}
\end{figure*}

\begin{figure*}
    \centering
    \includegraphics[width=\textwidth]{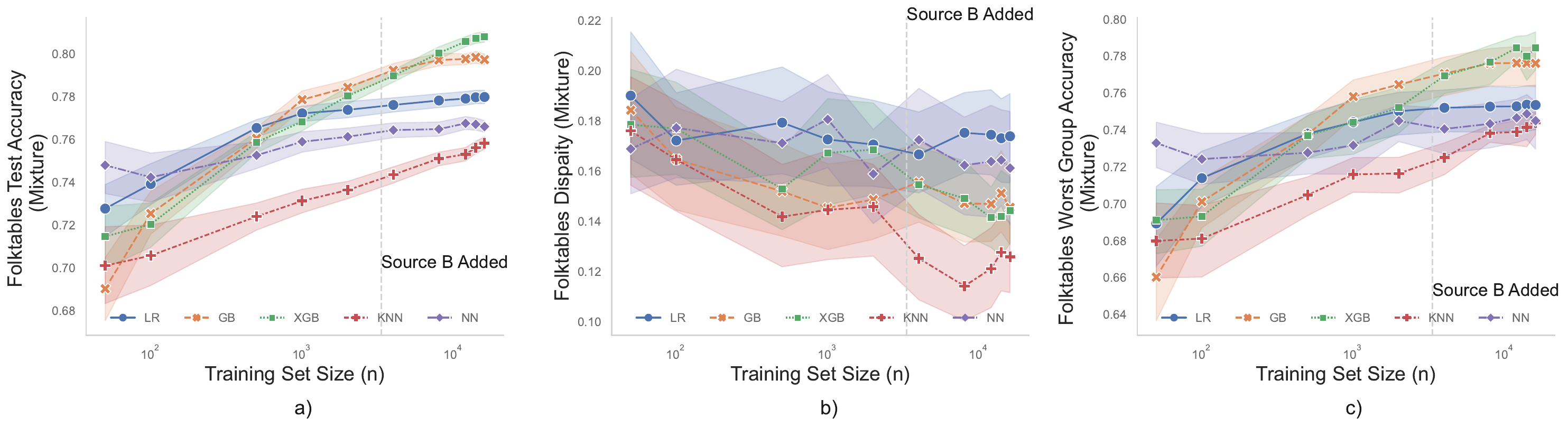}
    \caption{Folktables results on \textbf{Source A reference test set} in the \textsc{Mixture} case over 5 trials for \textbf{(a)} accuracy, \textbf{(b)}accuracy disparity, and \textbf{(c)} worst subgroup accuracy. The mixture is the same ratio as the final dataset for the \textsc{Sequential} case in Figure \ref{fig:folktables_seq}) (75\% CA and 25\% SD)}
    \label{fig:folktables_mix}
\end{figure*}

\paragraph{Single Source Datasets Benefit from Data Scaling Properties}
We consider the initial stage of the \textsc{Sequential} case---prior to sampling from any additional data sources---to be equivalent to a single-source data setting. We find that increasing the dataset size in this single source setting yields improved performance (Figure~\ref{fig:folktables_seq}a). 
Consistently, maximum accuracy is achieved when the most data points in a single source are used. 
Single source data increases also improve worst-subgroup performance, as demonstrated in prior literature~\citep{sagawa2019distributionally}. For some models, increases in data slightly improve disparity (e.g., XGB disparity drops from 17.0\% to 13.6\%)  while for other models even within the single source setting do little to minimize the differences between subgroup disparity\footnote{Results with confidence intervals and additional results in larger $n$ for single source scaling are available in the appendix}. 







\begin{figure*}
    \centering
    \includegraphics[width=0.9\textwidth]{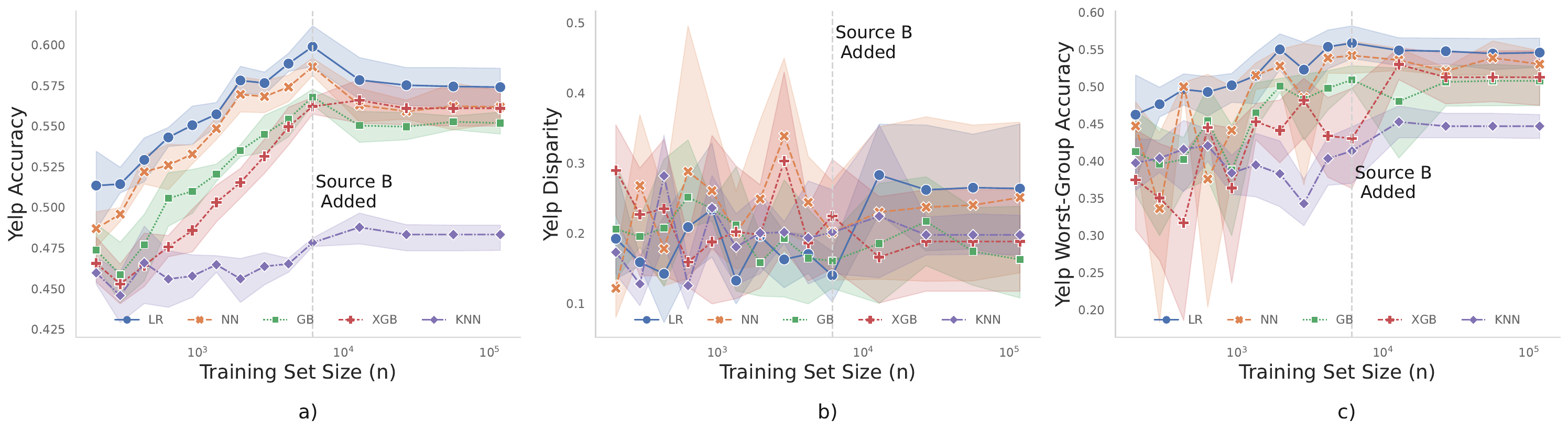}
    \caption{Yelp results on \textbf{Source A reference test set} in the \textsc{Sequential} case over 5 trials on  \textbf{(a)} accuracy, \textbf{(b)} accuracy disparity, and \textbf{(c)} worst subgroup accuracy. Source A is from New Jersey and Source B is from Pennsylvania.}
    \label{fig:yelp_dip}
\end{figure*}

\begin{figure*}
    \centering
    \includegraphics[width=0.9\textwidth]{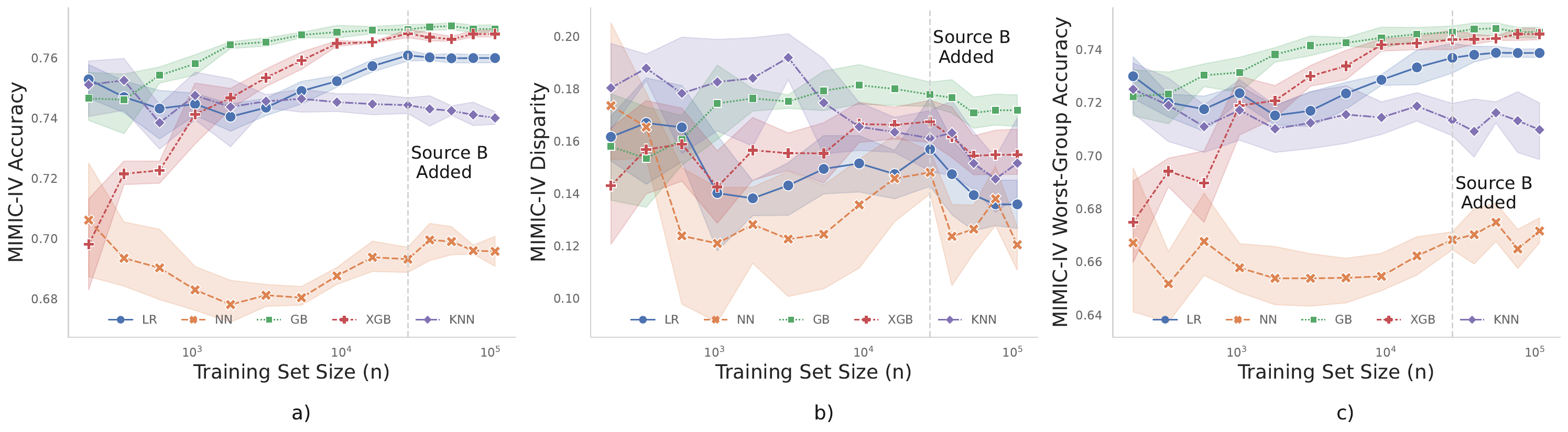}
    \caption{MIMIC-IV results on \textbf{Source A reference test set} in the \textsc{Sequential} case over 5 trials on  \textbf{(a)} accuracy, \textbf{(b)} accuracy disparity, and \textbf{(c)} worst subgroup accuracy. Source A is from admission type URGENT and Source B is from EW. EMER.}
    \label{fig:mimic_dip}
\end{figure*}

\paragraph{Multi-source Dataset Scaling Can Lead to Worse Outcomes on a Reference Test Set}

In the \textsc{Sequential} case (Figure ~\ref{fig:folktables_seq}a, \ref{fig:yelp_dip}a, \ref{fig:mimic_dip}a), we observe that adding additional data from a separate source, thereby quadrupling the size of the training set, leads to a dip in performance on a reference test set of interest, in addition to worse fairness metrics and worse robustness. For the Folktables dataset (Figure~\ref{fig:folktables_seq}a), this dip is observed empirically as a statistically significant reduction of test accuracy for all models except XGB (e.g., LR: -2.5\%; GB: -2.6\%; kNN: -3.5\%, NN: -2.7\%). This reduction in performance occurs when additional data is added from source B ($n_B$ = 12000) once source A ($n_A$ = 4000) is exhausted; the training data size has tripled. Decreases to worst subgroup performance were also significant and observed in all models except XGB (e.g., LR: -3.5\%; GB: -2.3\%; kNN: -2.3\%, NN: -2.7\%). We did not observe a significant decrease in disparity with the addition of more data. 

In the \textsc{Mixture} case (Figure ~\ref{fig:folktables_mix}), we find that scaling a fixed mixture of sources yields monotonically improved performance on the reference test set---i.e., there is no observed dip in performance, and the increase of the dataset size is correlated with increasing test accuracy, and better worst subgroup performance. From $n$ = 4000 to 16000, we see test accuracy consistently improve for all models (e.g., XGB:+1.8\%; kNN:+1.5\%) and we see worst subgroup performance also consistently improve across all models. As in the \textsc{Sequantial} case, we do not see a significant impact of data scaling on subgroup disparities due to the high variance in worst and best subgroup size in the test set. However, comparing the \textsc{Mixture} case to the \textsc{Sequential} over the same range of $n$ before all data is added, the accuracy achieved on the reference dataset for the best $n$ in the \textsc{Sequential} case remains consistently above the maximum accuracy in the \textsc{Mixture} case. At $n_{max}$ for \textsc{Sequential} case, test accuracy is higher (LR: 78.0\%, GB: 79.8\%, XGB: 80.8\%, kNN: 75.8\%, NN:76.7\%) than $n_{max}$ for the \textsc{Mixture} case (LR: 80.6\%, GB:82.3\%, XGB:81.0\%, kNN:79.4\%, NN: 79.6\%).

\paragraph{Multi-source Dataset Scaling Can Lead to Better Generalization}

\begin{figure}
    \centering
    \includegraphics[width=0.9\textwidth]{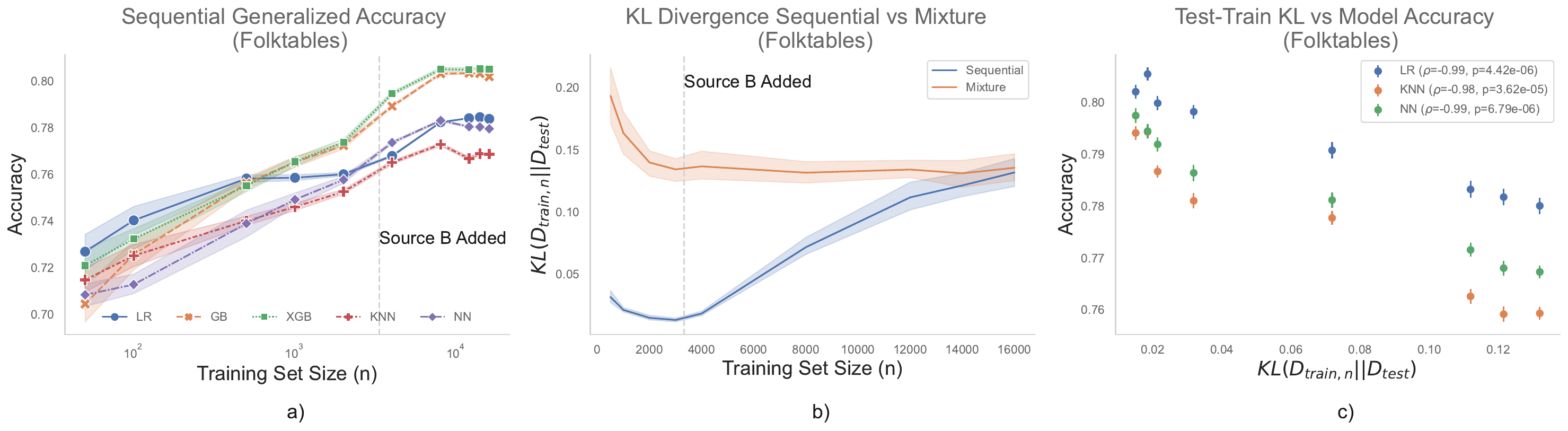}
    \caption{\textbf{(a)} Folktables accuracy in the \textsc{Sequential} case changes on the \textbf{generalized test set}, \textbf{(b)} Comparison of Train-Test KL: \textsc{Sequential} setting vs \textsc{Mixture} setting, \textbf{(c)} Relationship between Train-Test KL and Accuracy in the \textsc{Sequential} setting for different classifiers}
    \label{fig:gen}
\end{figure}

In (Figure~\ref{fig:yelp_dip}b, 
 \ref{fig:mimic_dip}b, \ref{fig:gen}a), we see that increases to the training dataset, even from sources of different distribution, still yield improvements on a generalized test set, sampled from all available sources (e.g., For $N$ = 50/4000/16000, LR: 72.7\% / 76.8\% / 78.4\%; GB: 70.4\% / 78.9\% / 80.2\%; XGB: 72.1\% / 79.5\% / 80.5\%; NN: 70.8\% / 77.4\% / 77.9\%).  This indicates that increasing dataset size across sources yields improvements to model generalization, even if this may not translate to improved performance on the reference test set of interest.




\subsection{Performance via the Lens of a Practical Divergence}
In the main text, the divergence measure we used was based on a score function because ICU data was too high-dimensional to apply density estimation techniques. For the Folktables dataset, we present the following results of computing KL divergence through kernel density estimation direction. 
\paragraph{Divergence comparison: \textsc{Sequential vs Mixture}} 
The first step is to empirically validate that our specific choice of divergence, KL divergence, increases in the \textsc{Sequential} setting as the training set size grows. We approximate densities through kernel density estimation with a Gaussian kernel on scaled PCA projections (3 components). Figure \ref{fig:gen}b compares the KL divergence between the training set and test set at different values of $n$ (data size) for the Folktables Income dataset. In the \textsc{Sequential} case, scaling up $n$ results in an increase in train-test divergence while in the \textsc{Mixture} case, this divergence remains static. 

\paragraph{Translating Divergence to Accuracy}
The next step is to validate that increased train-test divergence translates into a reduction in accuracy. We find a significant negative correlation between the KL divergence between the train and test dataset with the resulting model accuracy for 3 out of 5 models; as train-test divergence increases, test accuracy decreases. Figure \ref{fig:gen}c shows this correlation for the 3 algorithms where we observe a significant correlation. There was also a negative correlation between for Gradient Boosted Trees (GB). We did not observe a decrease in performance for XGBoost (XGB), thus such a correlation for the XGB model is not expected. 

\paragraph{Excess KL and Rejecting More Data}
\begin{figure*}
    \centering
    \includegraphics[width=\textwidth]{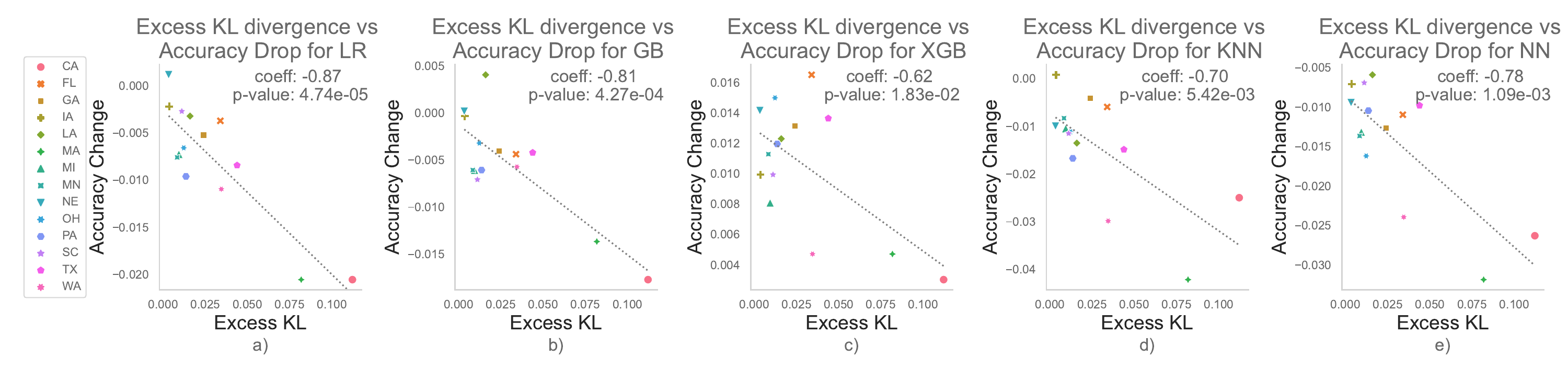}
    \caption{The relationship between Excess KL and the resulting accuracy drop from increasing for \textbf{a)} logistic regression (LR), \textbf{b)} gradient boosting (GB),  \textbf{c)} XGBoost (XGB),  \textbf{d)} K-Nearest Neighbor (kNN) and \textbf{e)} Neural Network (NN). We observe a statistically significant correlation between Excess KL and accuracy drop across all 5 models where we observe significant decreases in model performance.}
    \label{fig:kl}
\end{figure*}

Finally, we validate our proposed heuristic of excess KL ($\Delta_{KL}$) for deciding when to include more data (Section \ref{sec:lab}). If $\Delta_{KL}$ is larger than $0$ there is a significant distribution shift in the larger dataset and there is thus likely to be an increase or flat-lining of loss. We consider a large set of states as additional data sources: some are closer to South Dakota (e.g., Minnesota) while others are very distant (e.g., Florida). 

\begin{figure}
  \begin{center}
    \includegraphics[width=0.4\textwidth]{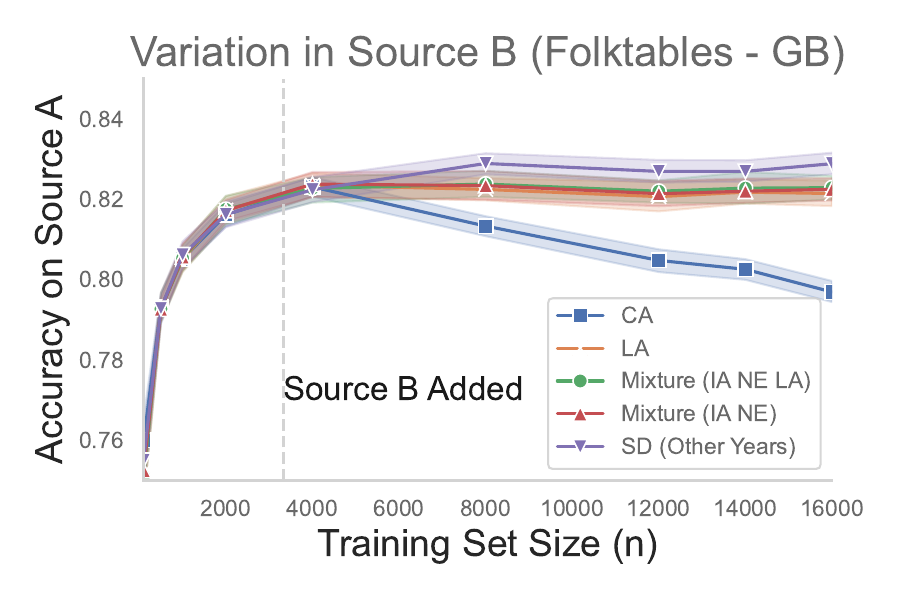}
      \caption{Accuracy on source A when different additional sources are used.}
  \end{center}
  \label{fig:sourceB}
\end{figure}

When comparing excess KL between the new bigger training distribution and the original dataset relative to the reference dataset, we find a significant negative correlation between accuracy change and excess KL across different states for all the classifiers. These results show that excess KL is indeed a reasonable heuristic for estimating the accuracy drop induced by additional data. Furthermore, the relative ordering of source states in terms of accuracy drop remains consistent across classifiers. 
However, the scale of accuracy drop for XGB is an order of magnitude smaller than other classifiers. Our results suggest that more data, albeit from a different distribution, does not affect XGB adversely to the same degree. 

If we replace California with a different state to be the additional data source based on excess KL, we observe better performance as more data (Figure~\ref{fig:sourceB}). Surprisingly, using Louisiana alone is as helpful as using a mixture of states near South Dakota (e.g., Nebraska and Iowa). Ultimately, the best improvement in performance comes from using South Dakota data from future years (2015-2018) but this source is only slightly better than using a state across the country (e.g. Louisiana) from the same year.


\subsection{Toy Experiments}
Before running real-world experiments, we first tested our concept on synthetic data. We consider two source A and source B where $y_A(x) = \sin(x)$ and $y_B(x) = -\sin(x)$. $\hat{D}_1$ is sized $n_A=10$ comes from source A  and $\hat{D}_2$ is sized $n_B=90$ comes from source B. The training set $\hat{D}_{train} = \hat{D}_1 \cup \hat{D}_2$ while the test set comes from just source A.

\begin{figure*}[h]
    \centering
    \includegraphics[width=\textwidth]{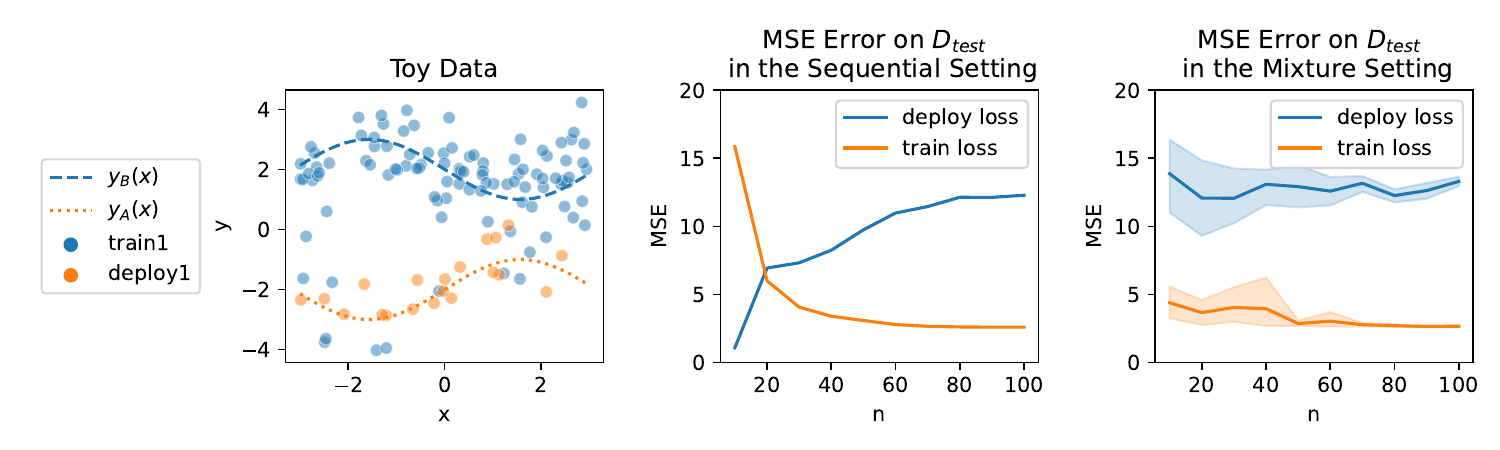}
    \caption{Toy experiments where in the sequential case, we see error on the test set getting worst as the size of the training examples in creases in the Sequential Setting}
    \label{fig:toy_learning}
\end{figure*}

\begin{figure*}
    \centering
    \includegraphics[width=\textwidth]{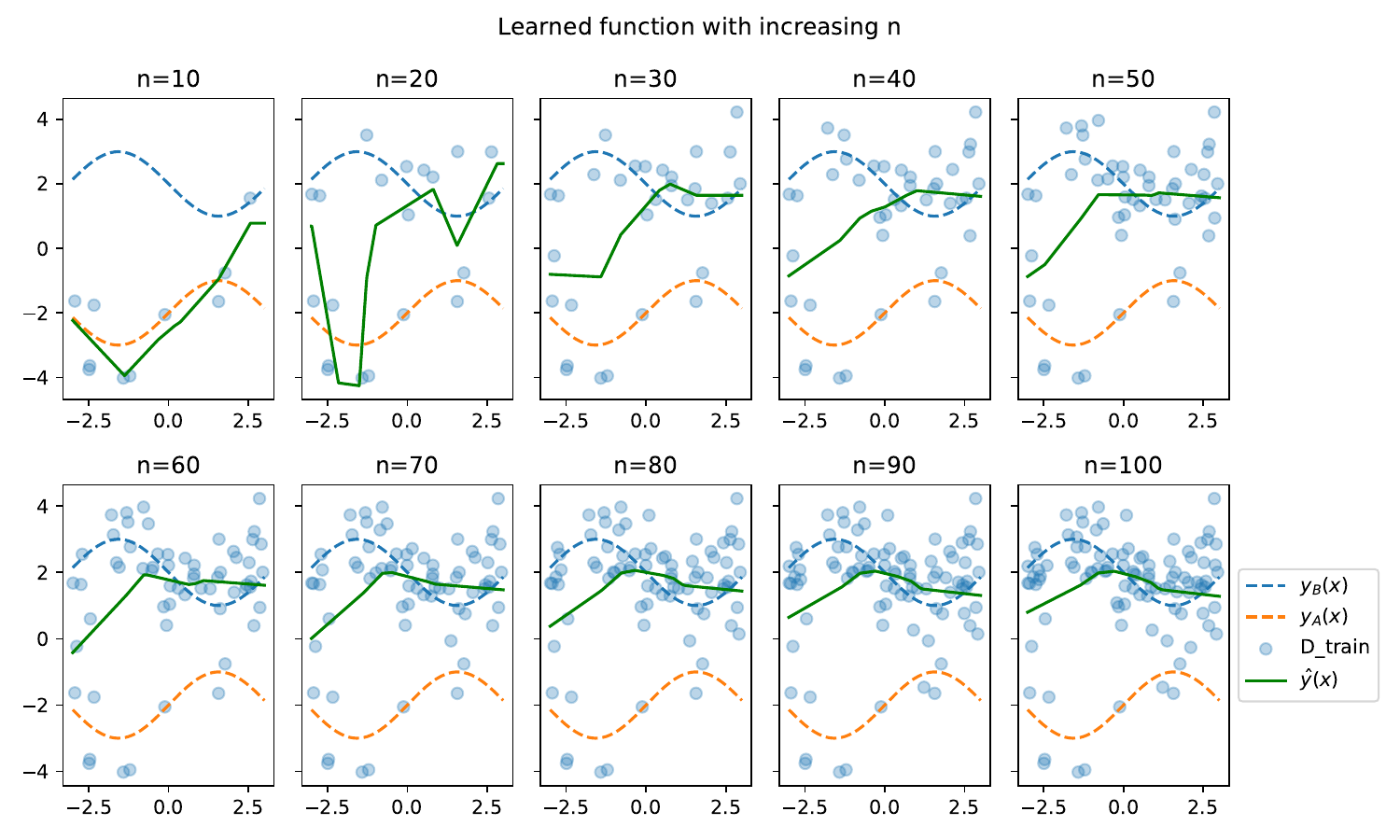}
    \caption{Visualization of learn function as the number of training data points increases. We see that $\hat{y}$ becomes closer to $y_B(x)$ as more data points are added from $\hat{D}_2$ in the sequential Setting.}
    \label{fig:toy_learning_time}
\end{figure*}

\section{Proofs}
\paragraph{Proof for Lemma 3.1}
\begin{lemma}

Let $D_{train, n}$ be constructed in the \textsc{Sequential} case from $k$ sources: $D_{S_1}, ..., D_{S_k}$, then if $\delta(D_{S_k}, D_{test}) - \frac{cn}{n_{s_k}} \ge \delta(D_{train, n}, D_{test}) $
\[
\delta(D_{train, n}, D_{test}) \ge \delta(D_{train, n-n_{s_k}}, D_{test})
\]
where $\delta$ belongs to the family of f-divergences and $c$ is a divergence-dependent constant where $ 
{\delta(D_{train,n}, D_{test}) + c = \sum_{i=1}^{m} \frac{n_{s_i}}{n} \delta(D_{S_i}, D_{test})}$.

\begin{proof}
Without loss of generality, the last source in the composition of $D_{train, n}$ is partially used, we define the size of the last source as simply $n_k$ and forget that there is unused data in the last source. Thus we can simply the overall training distribution as ${D_{train, n} = \sum_i^{k} \alpha_i D_{S_i}}$ where $\alpha_i=\frac{n_i}{n}$. Furthermore, let $n$ be the total number of examples with $k$ sources and let $n'=n-n_k$ be the total number of examples with $k-1$ sources. 
\begin{align*}
    \intertext{By convexity of $\delta$ (Jenson's):} 
    &\delta(D_{train, n'}, D_{test}) \\
    &\le \sum_i^{k-1} \frac{n_i}{n'}  \delta(D_{S_i}, D_{test})\\
    &= \sum_i^{k-1} \frac{n_i}{n'} \frac{n}{n}  \delta(D_{S_i}, D_{test}) \\
    &= \frac{n}{n'}   \sum_i^{k-1} \frac{n_i}{n} \delta(D_{S_i}, D_{test}) \\
    &= \frac{n}{n'} \big(\sum_i^{k} \frac{n_i}{n} \delta(D_{S_i}, D_{test}) - \frac{n_k}{n} \delta(D_{S_k}, D_{test}) \big) \\
    \intertext{Since $\delta(D_{train, n}, D_{test}) \le \sum_i^{k} \frac{n_i}{n} \delta(D_{S_i}, D_{test})$ then for some constant $c$:}
    &= \frac{n}{n'} \big( \delta(D_{train, n}, D_{test}) + c - \frac{n_k}{n} \delta(D_{S_k}, D_{test}) \big)\\ 
    \end{align*}
    \begin{align*}
    \intertext{Thus we need the following condition to be true:} \\
       \frac{n}{n'} \big( \delta(D_{train, n}, D_{test})  + \\ c - \frac{n_k}{n} \delta(D_{S_k}, D_{test}) \big) &\le \delta(D_{train, n}, D_{test}) \\
    \frac{n}{n'}\big( c - \frac{n_k}{n} \delta(D_{S_k}, D_{test}) \big) &\le (1 - \frac{n}{n'}) \delta(D_{train, n}, D_{test}) \\
    \frac{n}{n'}\big( c - \frac{n_k}{n} \delta(D_{S_k}, D_{test}) \big) &\le (\frac{n' - n}{n'}) \delta(D_{train, n}, D_{test}) \\ 
    n\big( c - \frac{n_k}{n} \delta(D_{S_k}, D_{test}) \big) &\le (n' - n) \delta(D_{train, n}, D_{test}) \\ 
    \frac{n}{n' - n}\big( c - \frac{n_k}{n} \delta(D_{S_k}, D_{test}) \big) &\ge \delta(D_{train, n}, D_{test}) \\ 
    \frac{n}{n-n'}\big(\frac{n_k}{n} \delta(D_{S_k}, D_{test}) - c\big) &\ge \delta(D_{train, n}, D_{test}) \\ 
    \frac{n_k}{n-n'} \delta(D_{S_k}, D_{test}) - \frac{cn}{n-n'} &\ge \delta(D_{train, n}, D_{test}) \\ 
    \intertext{since $n_k =n - n'$}
        \delta(D_{S_k}, D_{test}) - \frac{cn}{n_k} &\ge \delta(D_{train, n}, D_{test}) \\ 
\end{align*}
This final condition gives tells us that $\delta(D_{S_k}, D_{test})$ must be at least larger than $\delta(D_{train, n}, D_{test}$ in order for the resulting divergence after adding source $k$ to be larger. How much larger depends on $c$ which we can think of as the convexity constant of the divergence for f-divergences.
\end{proof}
\end{lemma}

\subsection{Connecting Sample Size, Divergence, and Performance}
Prior work has made the connection between source data and target data divergence and risk. We adapt the results to our notation as follows: 
\begin{theorem}{\citep{acuna2021f}} 
\label{thm:gen}
For $l:\mathcal{Y} \times\mathcal{Y} \rightarrow [0, 1] \in dom \phi*$,  every h in some hypothesis class $\mathcal{H}:$
\begin{align*}
    &\Err{l(h(x), y)}{(x, y) \sim \mathcal{D}_{test}} \le \Err{l(h(x), y)}{(x, y) \sim \mathcal{D}_{train, n}} +  \delta^{\phi}_{h, \mathcal{H}}(\mathcal{D}_{train, n}, \mathcal{D}_{test}) + \lambda
\end{align*}
where $\phi^*$ is the Fenchel conjugate of a convex, lower semi-continuous function $\phi$ that satisfies $\phi(1)=0$,  $\delta^{\phi}_{h, \mathcal{H}}$ is a discrepancy upper bounded by a corresponding f-Divergence, and $\lambda$ is the sum of risk from the ideal joint hypothesis $h^*$ over the train and test distributions (i.e. $\lambda = \Err{l(h^*(x), y)}{(x, y) \sim \mathcal{D}_{test}} + \Err{l(h^*(x), y)}{(x, y) \sim \mathcal{D}_{train, n}}$) 
\end{theorem}

\end{document}